\documentclass{article}

\PassOptionsToPackage{numbers, compress}{natbib}
\usepackage{natbib}
\usepackage{etoc}

 \usepackage[preprint]{neurips_2026}

\usepackage[utf8]{inputenc} 
\usepackage[T1]{fontenc}    
\usepackage{hyperref}       
\usepackage{url}            
\usepackage{booktabs}       
\usepackage{amsfonts}       
\usepackage{nicefrac}       
\usepackage{microtype}      
\usepackage{xcolor}         

\definecolor{mayablue}{rgb}{0.21,0.49,0.74}
\NewDocumentCommand{\hl}{ mO{} }{\textcolor{mayablue}{\textsuperscript{\textit{Hanlin}}\textsf{\textbf{\small[#1]}}}}
\NewDocumentCommand{\todo}{ mO{} }{\textcolor{mayablue}{\textsuperscript{\textit{Hanlin}}\textsf\textbf{\small[{TODO: #1]}}}}

\usepackage{amsmath,amssymb}
\usepackage{algorithm}
\usepackage{algpseudocode}
\usepackage{graphicx}
\usepackage{wrapfig}
\usepackage{enumitem}
\usepackage{cleveref}       
\usepackage{titlesec}
\titlespacing*{\paragraph}{0pt}{0pt}{0.5em}
\usepackage{caption}
\usepackage[most]{tcolorbox}

\definecolor{takeawayborder}{RGB}{106,81,163}   
\definecolor{takeawaybg}{RGB}{243,240,255}      

\newtcolorbox{takeawaybox}{
  colback=takeawaybg,
  colframe=takeawayborder,
  boxrule=0.9pt,
  arc=2pt,
  left=2pt,
  right=2pt,
  top=2pt,
  bottom=2pt
}

\title{
How Post-Training Shapes \\ Biological Reasoning Models
}

\author{%
  Lukas Fesser\thanks{Equal contribution.} \\
  Harvard University \\
  \texttt{lukas\_fesser@g.harvard.edu} \\ \vspace{-8mm} 
  \And
  Hanlin Zhang$^*$ \\
  Harvard University \\
  \texttt{hanlinzhang@g.harvard.edu} \\ \vspace{-8mm} 
  \And
  Michelle M. Li \\
  Harvard University \\
  \texttt{michelleli@g.harvard.edu} \\ \vspace{-8mm} 
  \AND
  Eric Wang \\
  Google DeepMind \\
  \texttt{ericzwang@google.com} \\ \vspace{-8mm} 
  \And
  Bryan Perozzi \\
  Google Research \\
  \texttt{bperozzi@google.com} \\ \vspace{-8mm} 
  \And
  Shekoofeh Azizi \\
  Google DeepMind \\
  \texttt{shekazizi@google.com} \\ \vspace{-8mm} 
  \AND
  Sham M. Kakade \\
  Harvard University \\
  \texttt{sham@seas.harvard.edu} \\ \vspace{-8mm} 
  \And
  Marinka Zitnik \\
  Harvard University \\
  \texttt{marinka@hms.harvard.edu} \\ \vspace{-8mm} 
}

\begin{document}

\etocdepthtag.toc{main}

\maketitle

\begin{abstract}
Scientific reasoning models for biology combine language models with foundation models trained on multimodal biological data, including DNA, RNA, and proteins. These models are built through post-training, yet how each stage shapes reasoning and generalization remains poorly understood.
We study when post-training improves performance and when it induces over-specialization. Across genomics, transcriptomics, and proteins, we train and evaluate more than 100 biological reasoning models under controlled variation in backbone, continued pre-training (CPT), supervised fine-tuning (SFT), and reinforcement learning (RL), measuring both in-domain (ID) and out-of-domain (OOD) performance.
We find that each post-training stage reshapes generalization in a distinct way rather than contributing uniform gains. CPT improves downstream performance by aligning models with biological language. SFT consistently increases ID performance but causes OOD performance to peak early and decline as models fit the training distribution. RL, when applied to strong SFT checkpoints with aligned rewards, improves OOD performance and partially recovers generalization.
These results show that biological reasoning does not improve monotonically with additional supervision or compute. Instead, performance depends on how training stages are composed. Under fixed post-training budgets, the strongest ID-OOD trade-off comes from brief SFT, larger RL allocations, and asymmetric adaptation capacity across stages. Code is available at \url{https://github.com/mims-harvard/bio-posttrain} and selected model checkpoints can be found here \url{https://huggingface.co/collections/mims-harvard/bio-posttrain}.

\end{abstract}
\section{Introduction}

Biology is becoming a central testbed for scientific reasoning models. Recent systems combine language models with biological foundation models trained on DNA, RNA, proteins, and other molecular data~\citep{fallahpour2025bioreason,istrate2025rbio1,fallahpour2026bioreason}. Their predictions require mapping natural-language task descriptions to molecular representations, integrating modality-specific evidence, and carrying intermediate biological state across multiple inference steps. Post-training is widely used to build such models, but its effects remain poorly understood across training stages~\citep{qievolm}. Despite strong empirical gains, it remains unclear how different stages of post-training shape reasoning and generalization.

Recent work has explored new forms of supervision, scaling strategies, and training objectives, including reinforcement learning for reasoning~\citep{guo2025deepseek,wang2025reinforcement,yue2025does}, large-scale post-training datasets~\citep{fan2025megascience,guha2025openthoughts}, and domain-specific adaptation pipelines~\citep{rizvi2026scaling,yu2026cdbridge}. Other studies examine how reward design~\citep{sun2025unleashing,zheng2025sci}, self-improving and world-model-based approaches~\citep{chencellduality,wei2025vcworld,su2026helix}, and training dynamics~\citep{chen2025reshaping,catalan2025training,chen2025coverage} influence model behavior. While these approaches improve task performance, they provide limited insight into how individual post-training stages affect generalization.

Biology provides a particularly stringent test of generalization. In mathematics and code, many out-of-domain problems retain the same underlying structure as the training examples, even when surface details change. In biology, unseen pathways, diseases, species, and perturbations often involve different mechanisms and biological processes~\citep{wei2026benchmarking,studer2026fully}. As a result, high benchmark performance does not necessarily indicate robust biological reasoning~\citep{durairaj2024plinder,queen2025procyon,ektefaie2024evaluating}. Models that perform well on familiar benchmarks may fail when transferred to new biological systems~\citep{kedzierska2025zero,ahlmann2025deep}. Additional post-training or larger models can therefore increase in-domain performance without improving biological generalization.

\begin{figure*}[t]
    \centering
    \includegraphics[width=\textwidth]{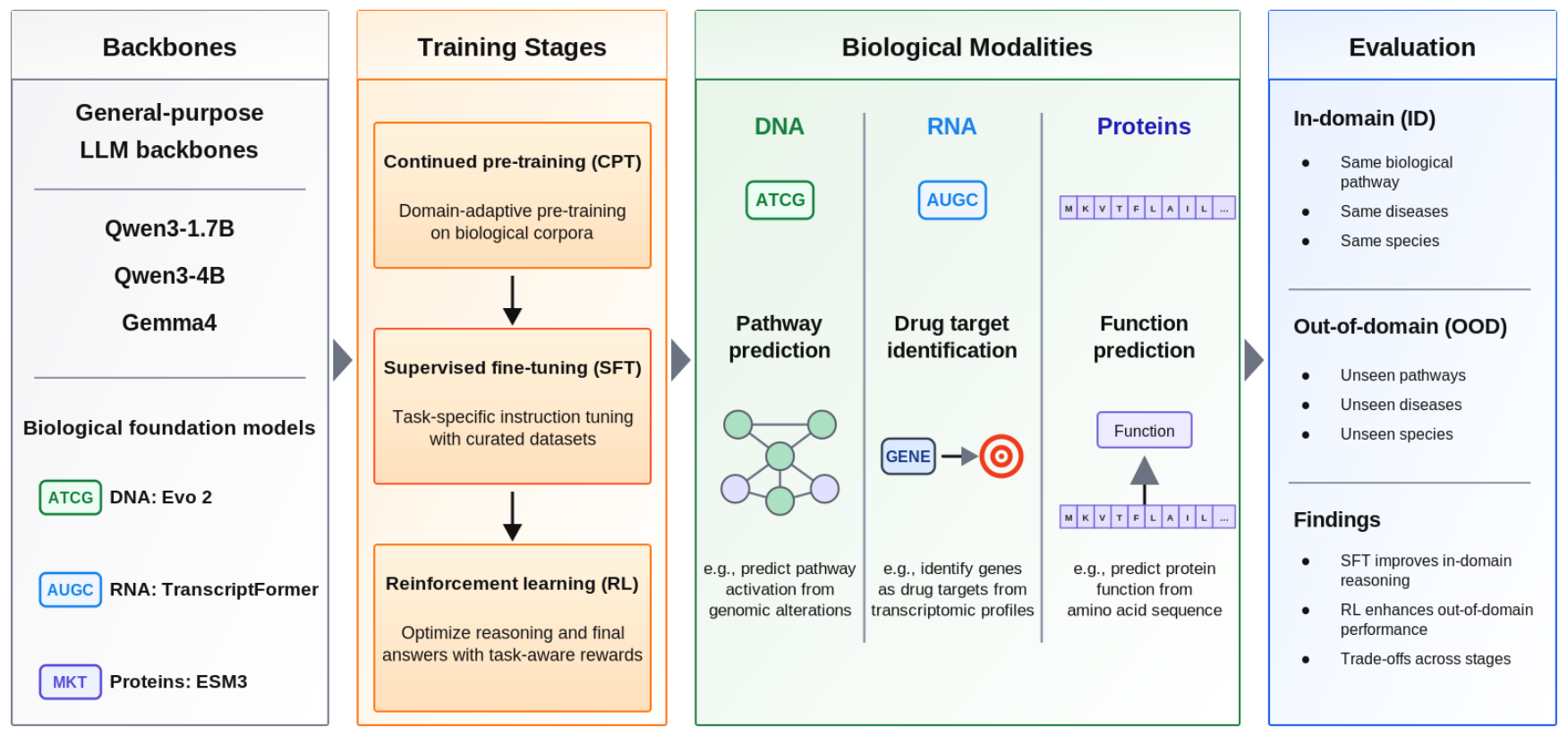}
    \caption{\textbf{Training dynamics define distinct generalization regimes in biological reasoning models.} We compare backbone choice, continued pre-training (CPT), supervised fine-tuning (SFT), and reinforcement learning (RL) across genomics, transcriptomics, and protein tasks, and evaluate each stage on biologically meaningful in-domain (ID) and out-of-domain (OOD) splits.}
    \vspace{-7mm}
    \label{fig:overview}
\end{figure*}

When additional post-training improves biological generalization, rather than primarily increasing fit to the training distribution, remains unclear. Existing studies typically examine one modality, one benchmark family, or one post-training stage at a time. Post-training itself is a sequence of stages, including continued pre-training, supervised fine-tuning, and reinforcement learning~\citep{wang2025loongrl,khatri2025art,jiangrethinking,feng2025cot,gao2026scpilot,tang2025ai,narayanan2025training}. These stages may interact in non-obvious ways, and gains from one stage may depend on the stages that precede or follow it. Yet existing models often differ simultaneously in backbone, data, scale, and supervision, making controlled comparisons difficult. Most studies also focus on final performance rather than training dynamics, and out-of-domain evaluation in biology remains limited and inconsistently defined.

\paragraph{Present work.}
We present a controlled study of post-training in biological reasoning models. Across genomics, transcriptomics, and proteins, we train and evaluate more than 100 biological reasoning models to examine when post-training improves biological generalization and when it primarily increases fit to the training distribution. Using matched model families, tasks, and data settings, we isolate the effects of backbone choice, continued pre-training (CPT), supervised fine-tuning (SFT), and reinforcement learning (RL) on both in-domain (ID) and out-of-domain (OOD) performance. We find that the same post-training budget can produce different generalization regimes depending on how it is allocated across stages: SFT increases ID performance but narrows OOD robustness, RL strengthens OOD performance when initialized from strong SFT checkpoints, and CPT improves downstream adaptation.

Our contributions are conceptual, empirical, and practical. Conceptually, we show that biological reasoning does not improve monotonically with additional post-training. Empirically, we present a controlled study across genomics, transcriptomics, and proteins that reveals consistent training dynamics across biological modalities. Practically, we derive design principles for post-training under limited compute. The strongest ID-OOD trade-off comes from combining brief SFT with larger RL allocations and allocating adaptation capacity asymmetrically across stages.

\section{Background and Related Work}

\paragraph{Biological Foundation Models and Multimodal Reasoning.}
Recent work has extended foundation modeling to a wide range of biological data~\citep{simon2024language}, including DNA~\citep{ji2021dnabert,zhou2023dnabert,nguyen2024sequence,brixi2026genome,dalla2025nucleotide,avsec2025alphagenome,cornman2024omg,hou2026phagebench}, RNA~\citep{fradkin2026orthrus,chen2022interpretable}, gene expression~\citep{pearce2025cross,rizvi2026scaling,cui2024scgpt,theodoris2023transfer,adduri2025predicting,hao2024large,istrate2024scgenept}, and proteins~\citep{lin2023evolutionary,hayes2025simulating,nijkamp2023progen2,alley2019unified}. Much of this literature focuses on learning representations or solving predictive tasks within a single modality, for example by modeling sequences or cellular profiles directly. More recent systems combine language models with biological inputs and structured context~\citep{istrate2025rbio1,fallahpour2025bioreason,queen2025procyon,schaefer2025multimodal}, enabling tasks that more closely resemble reasoning over pathways, cell states, or protein function.
Despite this progress, the literature remains fragmented. Models differ substantially in architecture, modality, training data, supervision format, and evaluation protocol, making it difficult to compare results or isolate which components drive performance. While recent biological foundation models provide a basis for multimodal reasoning, the field still lacks a unified view of how training choices affect downstream behavior, particularly under distribution shift.

\paragraph{Training Dynamics in LLMs: Training Stages Are Not Uniformly Additive.}
Modern language models are developed through a sequence of training stages rather than optimized in a single pass. A base model is first selected or pretrained, then adapted through domain-specific continued pre-training, task-specific supervised fine-tuning, and, in some cases, reinforcement learning~\citep{ouyang2022training}. These stages serve distinct roles. Continued pre-training expands coverage of domain vocabulary and structure while preserving general capabilities~\citep{que2024d}. Supervised fine-tuning specializes the model to task formats and target behaviors, often yielding large gains on benchmark tasks. Reinforcement learning further aligns outputs with task-dependent reward signals, improving performance but increasing sensitivity to reward design and sampling~\citep{kirk2024understanding}.

Recent work suggests that these stages are not uniformly additive~\citep{gururangan2020don,kaplan2020scaling,hoffmann2022training}. Performance gains can saturate or reverse as additional training is applied, and improvements on targeted evaluations need not translate to better generalization. \textsc{Pythia}~\citep{biderman2023pythia} enables fine-grained analysis of training dynamics across checkpoints and scales, while \textsc{EvoLM}~\citep{qievolm} characterizes non-monotonic behavior across training stages. Fine-tuning scaling laws further show that the interaction between model size, pretraining data, and supervision depends on the adaptation method~\citep{zhang2024scaling}. A growing body of work has studied how to conduct continued pre-training effectively: \citet{gupta2023continual} investigate learning rate re-warming schedules, \citet{ibrahim2024simple} propose scalable data mixing and replay strategies, \citet{parmar2024reuse} characterize when mid-training transfers general capabilities versus degrading them, and \citet{ke2023continual} introduce soft-masking to mitigate catastrophic forgetting. \citet{cheng2024adapting} further show that reformatting domain corpora as reading comprehension during continued pre-training yields stronger downstream performance than raw text exposure. Together with recent large-scale systems such as \textsc{Composer~2}~\citep{research2026composer2technicalreport}, these studies indicate that continued pre-training often serves as a critical transition stage enabling effective downstream adaptation~\citep{chu2025sft}, while supervised fine-tuning and reinforcement learning can produce strong task gains at the cost of reduced robustness if scaled or applied incautiously~\citep{wang2025gene,yin2025toward}.
\section{Experimental Setup: Tasks, Data, and Training Stages}

We design the experimental setup to isolate post-training effects while keeping the biological tasks, model families, and evaluation splits comparable across modalities. We first define the reasoning tasks and ID/OOD splits, then describe the model architecture and post-training pipeline.

\subsection{Biological Reasoning Tasks and Evaluation Splits}

We evaluate post-training across three domains: DNA, RNA, and proteins. Each task combines natural-language context with modality-specific biological inputs and uses ID and OOD splits. Dataset details and example prompts are provided in the appendix \ref{app:datasets}.

\paragraph{Pathway Prediction.} Pathway prediction asks the model to infer how a genetic variant propagates through a molecular pathway to produce a disease phenotype. We use the KEGG-derived benchmark introduced in \textsc{BioReason}~\citep{fallahpour2025bioreason}, which evaluates mechanistic reasoning over pathway structure rather than variant classification alone. Each example combines a reference DNA sequence, a variant DNA sequence, and textual pathway and gene context; the model generates a natural-language answer grounded in both sequence and pathway information. We define ID and OOD splits by pathway network, so OOD examples come from previously unseen molecular networks.

\paragraph{Drug Target Identification.} Target identification asks the model to choose the most promising therapeutic target for a disease and cell type. We adapt the cell-type-specific target nomination benchmark from \textsc{MEDEA}~\citep{sui2026medea}, simplifying it from an agentic tool-use setting to a fixed-input reasoning task. The model receives a natural-language description of the disease, cell type, and candidate genes, together with TranscriptFormer embeddings~\citep{pearce2025cross} for five candidates in normal and disease states, and identifies the best-supported target. We use four diseases for training and ID evaluation, and reserve hepatoblastoma as the OOD disease.

\paragraph{Protein Function Prediction.} Protein function prediction asks the model to infer the function of an uncharacterized protein from multimodal evidence. We build on the curated UniProt-based dataset introduced in \textsc{BioReason-Pro}~\citep{fallahpour2026bioreason}, which pairs experimentally supported GO annotations with protein-level context. Each example provides protein embeddings and text context, including organism, InterPro domain annotations~\citep{paysan2023interpro}, and protein--protein interactions. The model predicts protein function from this combined representation. We split data by species, with two held-out species forming the OOD test set, and remove the ontology graph inputs used in the original \textsc{BioReason-Pro} setup to align the task format with our study.


\subsection{Language Models, Biological Foundation Models, and Post-Training Pipeline}

We study biological reasoning through a common post-training pipeline built on general-purpose LLM backbones. Our main experiments use Qwen3-1.7B and Qwen3-4B, two dense models from the Qwen3 family, which support both reasoning-oriented and standard inference modes~\cite{yang2025qwen3}. We include Gemma 4 E2B as a backbone ablation to test whether the observed training dynamics persist under a different lightweight open model family~\cite{deepmind2026gemma4}. 

To represent biological modalities, we couple the LLM backbone to frozen biological foundation models through trainable projection layers. For DNA tasks, we use Evo2-1B, a genome foundation model trained for genome-scale sequence modeling and design across domains of life~\cite{brixi2026genome}. For RNA and transcriptomic tasks, we use TranscriptFormer, a cross-species single-cell model trained over evolutionary-scale transcriptomic data~\cite{pearce2025cross}. For protein tasks, we use ESM-3, a protein language model trained to model sequence, structure, and function across evolutionary scales~\cite{hayes2025simulating}. 


Post-training pipeline consists of three stages: (1) \textbf{Continued pre-training:} adaptation on general biological text using the standard next-token prediction loss. (2) \textbf{Supervised fine-tuning:} training on task-specific reasoning examples with the autoregressive language-model loss on target responses. (3) \textbf{Reinforcement learning:} optimization from a supervised checkpoint with a task-aligned reward objective, encouraging outputs that maximize task success rather than imitate reference traces. To enable controlled comparisons, we vary one factor at a time. In the main experiments, we scale SFT and RL compute, study the effect of CPT, and include ablations on backbone choice and LoRA rank. Full hyperparameters and implementation details are provided in the appendix.

We use a model signature to denote each configuration across training stages and biological domains. For example, {\color{mayablue}Q1-P-C-S$_{8, 20}$-R$_{16, 20}$} represents a model with the following setup:
\begin{itemize}[leftmargin=8pt, itemsep=0pt, topsep=0pt, parsep=0pt, partopsep=0pt]
    \item {\color{mayablue}Q1-P}: Qwen3-1.7B backbone evaluated in the protein setting. We use \texttt{D}, \texttt{R}, and \texttt{P} to denote DNA, RNA, and protein tasks, respectively. In our notation, \textcolor{mayablue}{blue} denotes Qwen3-1.7B models, \textcolor{orange}{orange} denotes Qwen3-4B models, and \textcolor{green!60!black}{green} denotes Gemma 4 E2B models.
    \item {\color{mayablue}C}: Continued pre-training on general biological text, specifically, the biology domain from FinefineWeb \citep{map2024finefineweb}.
    \item {\color{mayablue}S$_{8, 20}$}: Eight epochs of supervised fine-tuning on 20{,}000 task-specific reasoning traces.
    \item {\color{mayablue}R$_{16, 20}$}: Reinforcement learning for 16 epochs on 20{,}000 data points.
\end{itemize}
When a stage is omitted, the model has not undergone that part of the pipeline. For example, {\color{orange}Q4-R-S$_{4, 1}$} denotes a model fine-tuned directly from the Qwen3-4B + TranscriptFormer backbone  for 4 epochs on 1{,}000 reasoning traces in the RNA setting without CPT or RL, while {\color{green!60!black}G-R-C-S$_{8, 1}$} denotes a Gemma + TranscriptFormer model adapted with CPT and then fine-tuned for 8 epochs.

\section{Results: How Training Stages Shape Biological Reasoning in LLMs}

We now present our main results related to scaling post-training for biology in compute- or data-bound settings. Concrete model settings for all our experiments in this section, including context windows, input sequence lengths, and other hyperparameters can be found in the appendix.

\begin{figure*}[h!]
    \centering
    \includegraphics[width=\textwidth]{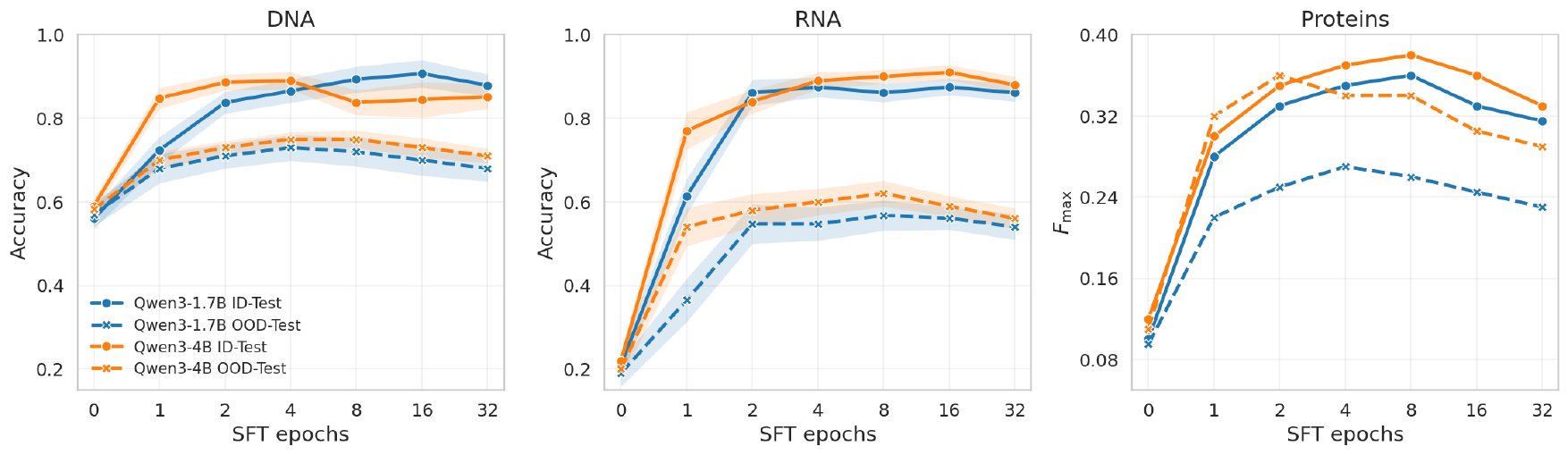}
    \caption{\textbf{Supervised fine-tuning improves in-domain performance but reduces out-of-domain robustness.} As SFT compute increases, ID performance continues to improve, while OOD performance peaks early and declines, indicating over-specialization to the training data. DNA/ RNA mean and std. over 3 random seeds (we only use one seed for Proteins, due to the size of the dataset).
    }
    \label{fig:sft_epoch_scaling}
\end{figure*}

\subsection{Supervised Fine-Tuning Increases Accuracy but Narrows Generalization}

We begin by studying how supervised fine-tuning scales in biological reasoning models using pretrained Qwen3-1.7B and Qwen3-4B backbones~\citep{yang2025qwen3}.

\paragraph{Fixed data, variable compute.} We first consider a data-constrained regime based on the DNA and RNA tasks. For each backbone, we train model families of the form {\color{mayablue}\textsc{Q1-D-S}$_{\{1,2,4,8,16,32\},1}$}, {\color{orange}\textsc{Q4-D-S}$_{\{1,2,4,8,16,32\},1}$}, {\color{mayablue}\textsc{Q1-R-S}$_{\{1,2,4,8,16,32\},1}$}, and {\color{orange}\textsc{Q4-R-S}$_{\{1,2,4,8,16,32\},1}$}, where the subscript indicates the number of SFT epochs and the use of the full available training set in each domain. We then evaluate both in-domain and out-of-domain performance. 
Figure~\ref{fig:sft_epoch_scaling} reveals a generalization trade-off induced by supervised fine-tuning. The amount of training that maximizes ID performance is consistently larger than the amount that maximizes OOD performance, indicating that continued fine-tuning improves fit to the training distribution after OOD generalization has already peaked.
In DNA, for example, {\color{mayablue}\textsc{Q1-D-S}} improves its ID accuracy from roughly \(0.68\) at 1 epoch to about \(0.90\) by 16 epochs, while its OOD accuracy peaks much earlier, around \(0.73\) at 2--4 epochs, and then declines to about \(0.68\) by 32 epochs. The same pattern appears for the larger {\color{orange}\textsc{Q4-D-S}} model.

The RNA setting shows the same trend. {\color{mayablue}\textsc{Q1-R-S}} gains nearly \(0.3\) ID accuracy points from 1 to 4 epochs, but while its OOD performance also improves, it comes within at most \(0.3\) of ID accuracy. For {\color{orange}\textsc{Q4-R-S}}, ID accuracy rises from roughly \(0.78\) at 1 epoch to about \(0.91\) by 16 epochs, whereas OOD accuracy peaks much earlier, around 4 epochs, and then drifts downward by the end of training. This same pattern persists when more supervision is available: on the proteins task, {\color{mayablue}\textsc{Q1-P-S}$_{\{1,2,4,8,16,32\},20}$} and {\color{orange}\textsc{Q4-P-S}$_{\{1,2,4,8,16,32\},20}$} both continue to improve ID \(F_{\max}\) through about 8 epochs, but OOD \(F_{\max}\) peaks earlier and then declines. More SFT compute therefore improves in-domain accuracy more reliably than it improves biological generalization.

\begin{wrapfigure}{r}{0.52\textwidth}
    \centering
    \includegraphics[width=0.50\textwidth]{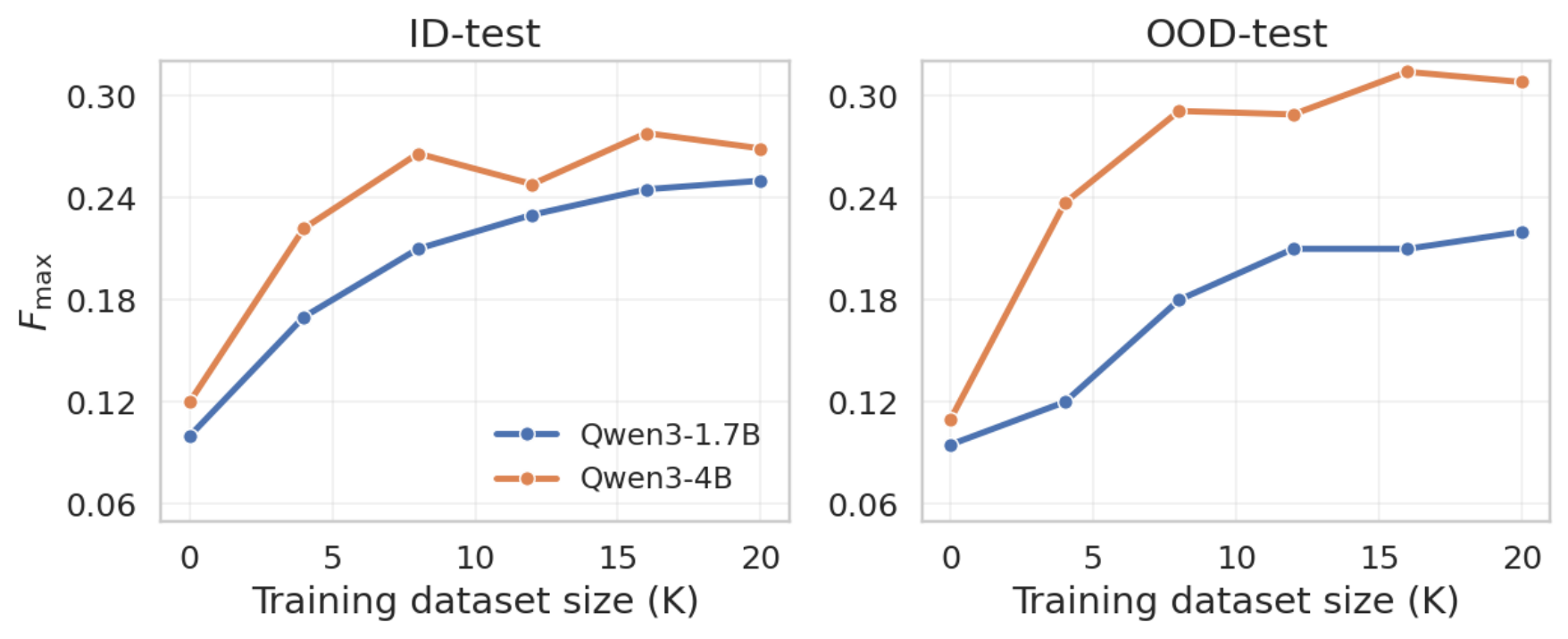}
    \caption{\textbf{Increasing data improves generalization more reliably than increasing SFT epochs.} Scaling dataset size yields gains in both ID and OOD performance, but with diminishing returns, in contrast to the overfitting behavior observed when scaling epochs.
    }
    \label{fig:sft_data_scaling}
\end{wrapfigure}

\paragraph{Fixed compute, variable data.} We next study a compute-constrained regime in proteins by fixing training to a single SFT epoch and varying the amount of supervision. Concretely, we train {\color{mayablue}\textsc{Q1-P-S}$_{1,\{4, 8, 12, 16, 20\}}$} and {\color{orange}\textsc{Q4-P-S}$_{1,\{4, 8, 12, 16, 20\}}$}. Figure~\ref{fig:sft_data_scaling} shows that increasing data is better behaved than increasing epochs on a fixed dataset. For {\color{mayablue}\textsc{Q1-P-S}}, scaling from 4K to 20K training examples increases ID \(F_{\max}\) from \(0.17\) to \(0.23\) and OOD \(F_{\max}\) from \(0.11\) to \(0.22\). For {\color{orange}\textsc{Q4-P-S}}, the corresponding gains are from about \(0.22\) to \(0.27\) on ID and from about \(0.24\) to just above \(0.30\) on OOD. While these gains are significant, they flatten quickly: after roughly 40--60\% of the data, both curves improve only marginally.

These results suggest that SFT is a strong driver of in-domain biological reasoning, but that scaling it naively, either through more epochs or more data, does not reliably translate into better OOD performance. Instead, the dominant pattern is over-specialization: the model becomes better at the benchmark distribution while becoming less robust to biological shift. At the same time, the protein results indicate that these two ways of scaling compute are not equivalent. Increasing epochs on a fixed dataset produces the sharper trade-off, with OOD performance peaking early and then declining as the model repeatedly fits the same supervision. Increasing data while holding epochs fixed produces a more stable pattern. ID performance still shows diminishing returns, but OOD performance remains roughly monotonic and then plateaus rather than collapsing. Under a fixed compute budget, data scaling is therefore the more robust strategy when additional supervision is available.

\begin{takeawaybox}
\textbf{Finding 1.} Supervised fine-tuning rapidly increases in-domain performance but progressively concentrates models on the training distribution, reducing robustness under biological distribution shift.
\end{takeawaybox}

\subsection{Reinforcement Learning Recovers Generalization After Fine-Tuning}

\begin{figure*}[h!]
    \centering
    \includegraphics[width=\textwidth]{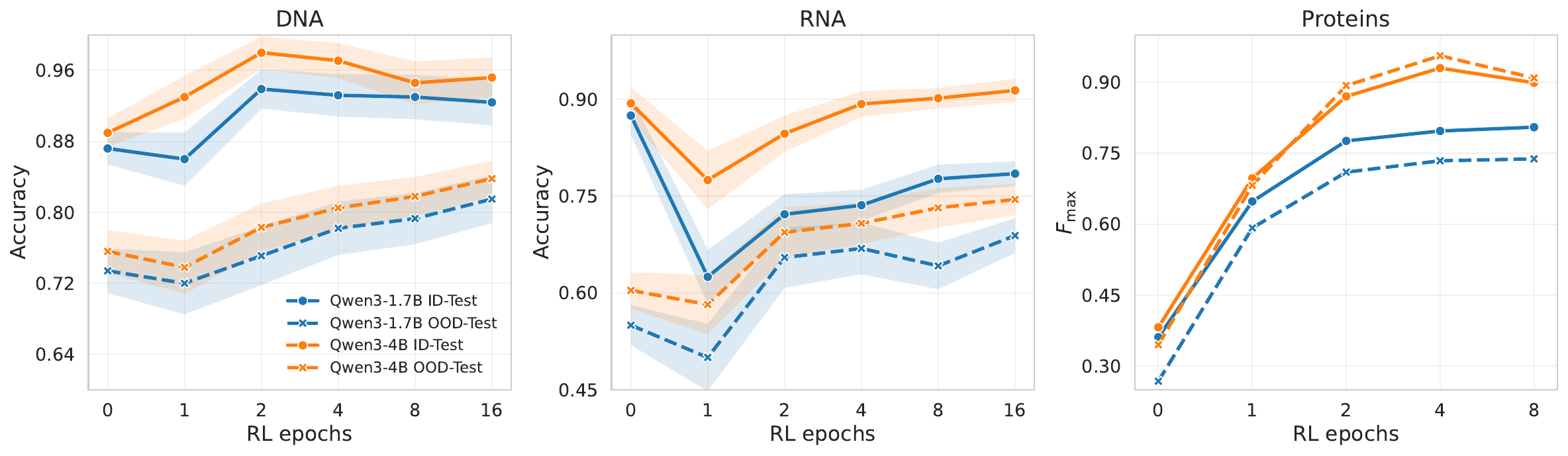}
    \caption{\textbf{Reinforcement learning consistently improves out-of-domain robustness.} Starting from strong SFT checkpoints, RL increases both ID and OOD performance, with the largest gains in OOD and diminishing returns after the first few epochs. DNA/ RNA mean and std. over 3 random seeds (we only use one seed for Proteins, due to the size of the dataset).
    }
    \label{fig:rl_epoch_scaling}
\end{figure*}

\paragraph{Scaling RL epochs.} We now ask whether reinforcement learning can recover some of the robustness lost under SFT. Starting from the strongest SFT checkpoints selected on validation performance, we train model families of the form {\color{mayablue}\textsc{Q1-D-S}$_{8,1}$}{\color{mayablue}\textsc{-R}$_{\{1,2,4,8,16\},1}$},
{\color{orange}\textsc{Q4-D-S}$_{4,1}$}{\color{orange}\textsc{-R}$_{\{1,2,4,8,16\},1}$}, {\color{mayablue}\textsc{Q1-R-S}$_{4,1}$}{\color{mayablue}\textsc{-R}$_{\{1,2,4,8,16\},1}$}, and {\color{orange}\textsc{Q4-R-S}$_{8,1}$}{\color{orange}\textsc{-R}$_{\{1,2,4,8,16\},1}$} in the DNA and RNA settings, together with {\color{mayablue}\textsc{Q1-P-S}$_{4,20}$}{\color{mayablue}\textsc{-R}$_{\{1,2,4,8,16\},20}$} and {\color{orange}\textsc{Q4-P-S}$_{4,20}$}{\color{orange}\textsc{-R}$_{\{1,2,4,8,16\},20}$} in the protein setting. Figure~\ref{fig:rl_epoch_scaling} shows that, unlike SFT, RL improves both ID and OOD performance quite consistently over the range we study. The gains are not only directional but substantial: in DNA, OOD accuracy rises by about \(0.05\) over the RL sweep, and in proteins the OOD improvement is larger still, especially for {\color{orange}\textsc{Q4-P-S}$_{4,20}$}{\color{orange}\textsc{-R}$_{\{1,2,4,8,16\},20}$}, which gains roughly \(0.08\) absolute \(F_{\max}\) from the first to the best RL checkpoint. Across tasks, the largest gains typically appear in the first few RL epochs, with later epochs yielding smaller additional improvements, especially OOD.



\begin{takeawaybox}
\textbf{Finding 2.} Starting from a strong SFT checkpoint, reinforcement learning shifts models toward a stronger ID-OOD frontier. The largest gains occur out-of-domain, and most of the improvement appears within the first few RL epochs.
\end{takeawaybox}

\subsection{Continued Pre-Training Establishes the Foundation for Biological Reasoning}

\begin{figure}[h]
    \centering
    \includegraphics[width=\textwidth]{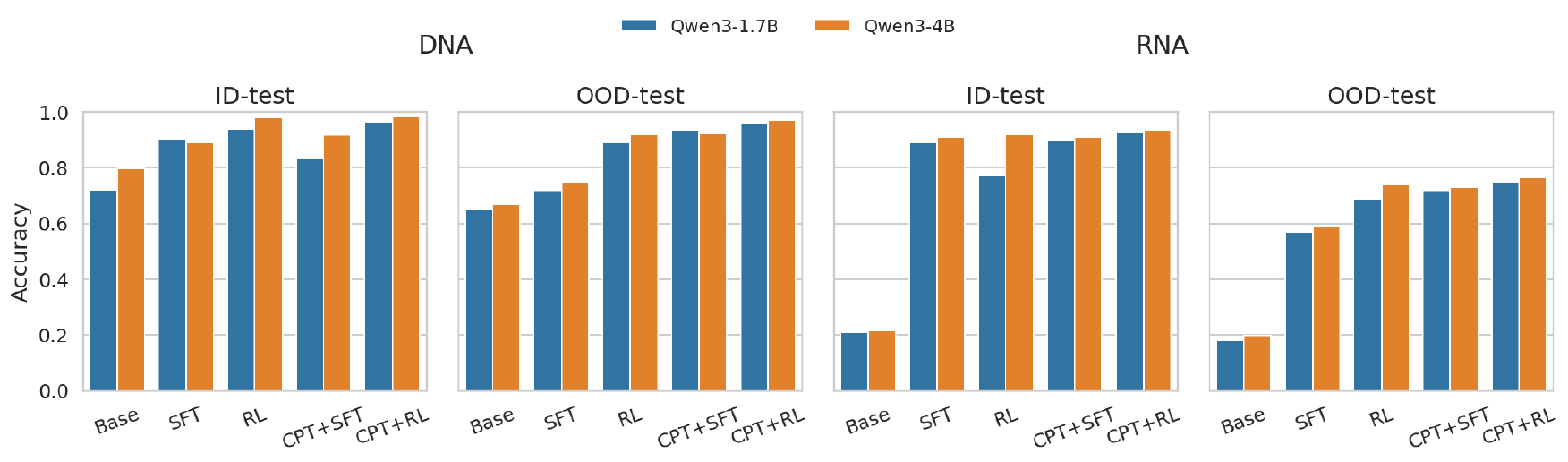}
    \caption{\textbf{Continued pre-training improves the effectiveness of downstream post-training.} CPT improves both SFT and RL performance, with the largest gains appearing after RL and in out-of-domain settings.
    }
    \label{fig:cpt_effects}
\end{figure}

We next study whether continued pre-training changes how much downstream post-training can help. In the DNA and RNA settings, we first adapt the base backbones with continued pre-training on biological texts, yielding model families of the form {\color{mayablue}Q1-D/R-C} and {\color{orange}Q4-D/R-C}. We then evaluate these models under the strongest post-training configurations identified above, namely {\color{mayablue}Q1-D-C-S}$_{\color{mayablue}8,1}$, {\color{orange}Q4-D-C-S}$_{\color{orange}4,1}$, {\color{mayablue}Q1-D-C-S}$_{\color{mayablue}8,1}${\color{mayablue}-R}$_{\color{mayablue}16,1}$, and {\color{orange}Q4-D-C-S}$_{\color{orange}4,1}${\color{orange}-R}$_{\color{orange}16,1}$ for DNA and {\color{mayablue}Q1-R-C-S}$_{\color{mayablue}4,1}$, {\color{mayablue}Q1-R-C-S}$_{\color{mayablue}4,1}${\color{mayablue}-R}$_{\color{mayablue}8,1}$, 
{\color{orange}Q4-R-C-S}$_{\color{orange}8,1}$, and {\color{orange}Q4-R-C-S}$_{\color{orange}8,1}${\color{orange}-R}$_{\color{orange}8,1}$ for RNA. Figure~\ref{fig:cpt_effects} shows that CPT improves downstream performance at almost every stage we test, but that the size of the gain depends strongly on the stage. The gains are modest at the SFT stage in domain and markedly larger after RL, especially out-of-domain. Here, CPT lifts ID and OOD performance by visibly larger margins than SFT alone.

This effect is especially pronounced for the smaller {\color{mayablue}Q1-D} model out-of-domain, where CPT improves the effectiveness of SFT and RL by 0.2 and 0.08, respectively. For {\color{orange}Q4-D}, the same pattern holds, but from a stronger starting point and with smaller absolute gains. These findings qualitatively also hold in the RNA setting and suggest that especially for smaller models, CPT can act as a bridge between a general-purpose backbone and the reasoning demands of biology. 
Without CPT, downstream training must learn biological language, task structure, and reasoning behavior at the same time.

\begin{takeawaybox}
\textbf{Finding 3.} Continued pre-training, even on general biological texts, improves the effectiveness of both SFT and RL, with the strongest gains appearing OOD when post-training is applied after the model has first adapted to biological language.
\end{takeawaybox}

\subsection{Backbone Strength Shifts Performance Ceiling but Not Training Dynamics}

\begin{wrapfigure}{r}{0.58\textwidth}
    \centering
    \includegraphics[width=0.56\textwidth]{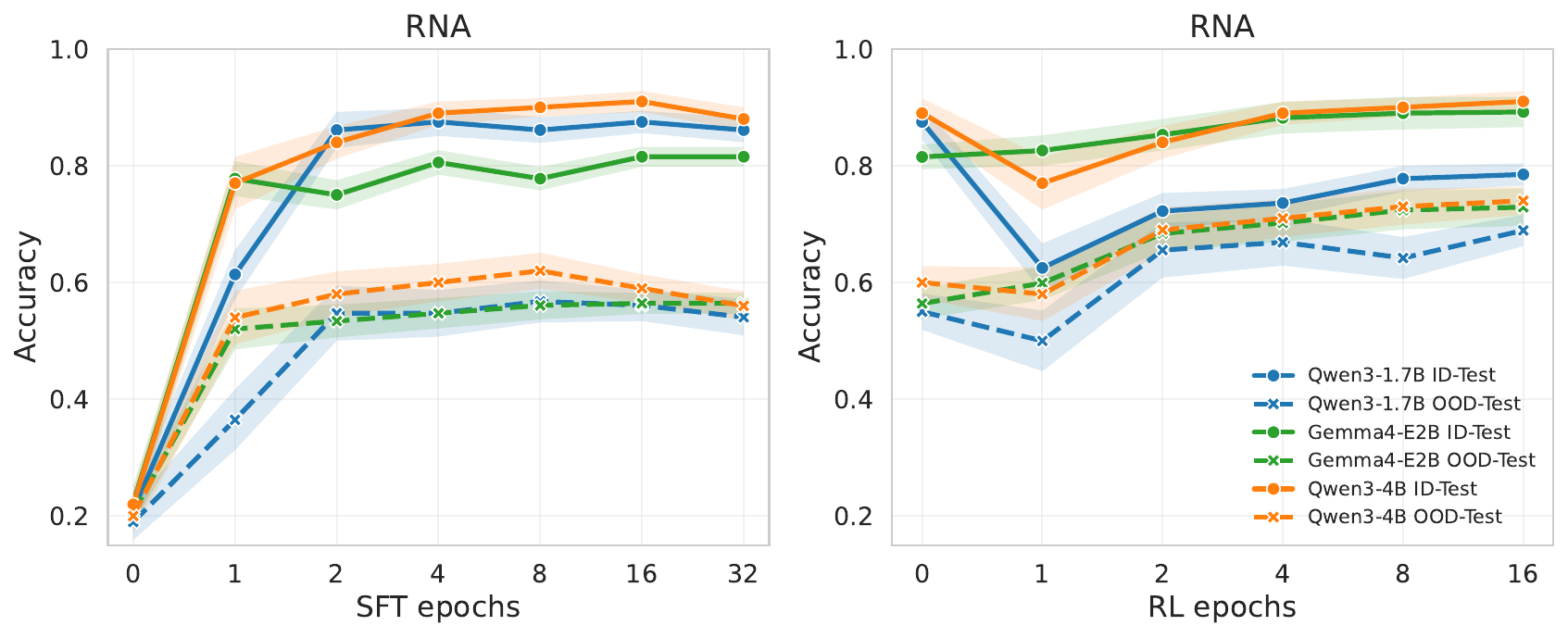}
    \caption{\textbf{Stronger backbones improve performance achievable with post-training but preserve training dynamics.} {\color{green!60!black}G-R} does not display an initial drop in performance when starting RL, unlike {\color{mayablue}Q1-R} and generally performs better OOD. Mean and std. over 3 random seeds.}
    \label{fig:gemma}
\end{wrapfigure}

To test whether our main findings depend on the choice of base model, we repeat the RNA experiments with an off-the-shelf backbone Gemma model ~\citep{deepmind2026gemma4}. In addition to {\color{mayablue}Q1-R} and {\color{orange}Q4-R}, we evaluate the more recent Gemma4-E2B backbone, denoted {\color{green!60!black}G-R}. Concretely, for SFT scaling we train model families of the form {\color{mayablue}Q1-R-S}$_{\color{mayablue}\{1,2,4,8,16,32\},1}$, {\color{orange}Q4-R-S}$_{\color{orange}\{1,2,4,8,16,32\},1}$, and {\color{green!60!black}G-R-S}$_{\color{green!60!black}\{1,2,4,8,16,32\},1}$. For RL scaling, we then start from the strongest SFT checkpoint (as measured by validation loss) for each backbone and train {\color{mayablue}Q1-R-S}$_{\color{mayablue}4,1}${\color{mayablue}-R}$_{\color{mayablue}\{1,2,4,8,16\},1}$, {\color{orange}Q4-R-S}$_{\color{orange}8,1}${\color{orange}-R}$_{\color{orange}\{1,2,4,8,16\},1}$, and {\color{green!60!black}G-R-S}$_{\color{green!60!black}4,1}${\color{green!60!black}-R}$_{\color{green!60!black}\{1,2,4,8,16\},1}$. With SFT only, {\color{green!60!black}G-R} trails the smaller {\color{mayablue}Q1-R} models OOD, but are somewhat weaker in-domain, as Figure~\ref{fig:gemma} shows.

After one epoch, {\color{green!60!black}G-R} is still comparable to {\color{orange}Q4-R} and outperforms {\color{mayablue}Q1-R} by around \(0.17\) both in- and out-of-domain. After that, more SFT helps the Qwen models more, at least in-domain, but {\color{green!60!black}G-R} accuracy still increases by \(0.03\). The qualitative trend is therefore unchanged: supervised fine-tuning is most effective for in-domain performance, but out-of-domain performance peaks substantially earlier and then plateaus or even declines with additional epochs. Backbone quality therefore shifts the overall SFT frontier upward, but does not remove the core ID-OOD trade-off induced by SFT.

Reinforcement learning exhibits a similar pattern. Starting from the strongest SFT checkpoint for each backbone, Figure~\ref{fig:gemma} shows that RL improves OOD performance more reliably than additional SFT, and that the larger backbones benefit more smoothly from this stage. In particular, {\color{green!60!black}G-R-S}$_{\color{green!60!black}4,1}${\color{green!60!black}-R}$_{\color{green!60!black}\{1,2,4,8,16\},1}$ qualitatively follows the larger {\color{orange}Q4-R-S}$_{\color{orange}8,1}${\color{orange}-R}$_{\color{orange}\{1,2,4,8,16\},1}$ trajectory more closely than the smaller {\color{mayablue}Q1-R-S}$_{\color{mayablue}4,1}${\color{mayablue}-R}$_{\color{mayablue}\{1,2,4,8,16\},1}$ model. Both {\color{green!60!black}G-R} and {\color{orange}Q4-R} improve more steadily under RL and do not show the initial drop visible for {\color{mayablue}Q1-R} when RL begins. Instead, Gemma improves monotonically by 0.08 ID and 0.15 OOD between RL epochs 1 and 16. This seems to indicate that backbone choice does not qualitatively alter the role of the training stage itself. The backbone matters primarily for the level of performance achievable with post-training, whereas the structure of the training dynamics appears to be stable across model families.

\begin{takeawaybox}
\textbf{Finding 4.} Larger models/ stronger backbones can raise absolute performance and tend to make RL gains more reliable for biological reasoning, but they do not change the basic training dynamics, including the SFT-induced ID-OOD trade-off.
\end{takeawaybox}

\subsection{Adaptation Capacity Should Be Allocated Asymmetrically Between SFT and RL}

\begin{figure*}[h!]
    \centering
    \includegraphics[width=\textwidth]{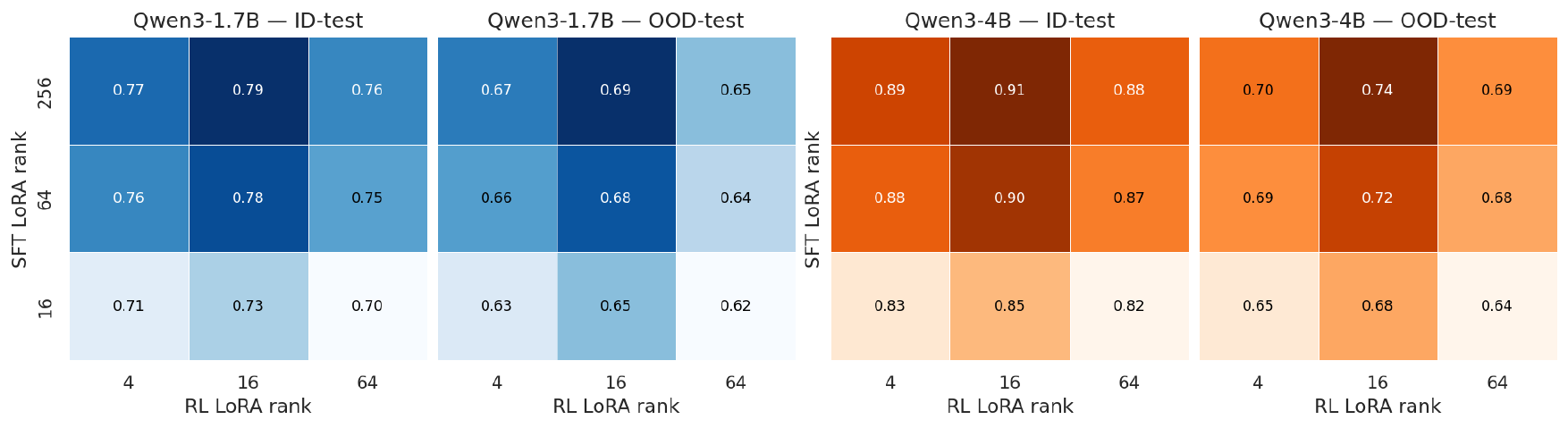}
    \caption{\textbf{Optimal adaptation requires asymmetric capacity across training stages.} Higher LoRA rank benefits SFT, while lower rank is sufficient for RL, indicating that different stages require different adaptation capacity (both for ID and OOD tasks). Shown are results for drug target identification (RNA) tasks.
    }
    \label{fig:sft_scaling}
\end{figure*}

We further study how adaptation capacity should be allocated across post-training stages by running a joint SFT--RL LoRA ablation in the RNA setting. Using {\color{mayablue}Q1-R-S}$_{\color{mayablue}4,1}${\color{mayablue}-R}$_{\color{mayablue}8,1}$ and {\color{orange}Q4-R-S}$_{\color{orange}8,1}${\color{orange}-R}$_{\color{orange}8,1}$ as our reference model families, we vary the SFT LoRA rank over \(r_{\mathrm{SFT}} \in \{16,64,256\}\) and the RL LoRA rank over \(r_{\mathrm{RL}} \in \{4,16,64\}\), with the corresponding scaling factors set proportionally to rank. For each backbone, every model is first fine-tuned with SFT using the same training data, optimizer, and epoch budget as in our main RNA experiments, and is then further optimized with RL using the same reward and training schedule. We evaluate the final checkpoint from each configuration on both ID-test and OOD-test splits, and summarize the results as heatmaps over \((r_{\mathrm{SFT}}, r_{\mathrm{RL}})\). This setup isolates whether the best end-to-end pipeline prefers symmetric adapter budgets across stages or an asymmetric allocation in which SFT and RL use different amounts of trainable capacity.

We find a clear asymmetry between the two stages. In both backbones, increasing the SFT rank from \(16\) to \(64\) or \(256\) produces a clear upward shift in ID performance and usually also improves OOD performance, whereas increasing the RL rank beyond \(16\) yields much smaller gains and can even reduce OOD performance. The highest ID regions in Figure~\ref{fig:sft_scaling} cluster at \(r_{\mathrm{SFT}}=256\), showing that SFT benefits from having enough capacity to absorb task format, domain structure, and multimodal reasoning patterns. By contrast, for RL, the strongest ID and OOD regions are concentrated at \(r_{\mathrm{RL}}=16\). The best overall configurations are therefore not those with matched ranks, but those with \textbf{high-capacity SFT and low-capacity RL}. This pattern holds for both {\color{mayablue}Q1-R} and {\color{orange}Q4-R}, suggesting that post-training in biological reasoning should be stage-specific not only in compute and data allocation, but also in adaptation capacity.

\begin{takeawaybox}
\textbf{Finding 5.} The best post-training pipelines use asymmetric adaptation capacity, with larger LoRA ranks for SFT and smaller ones for RL.
\end{takeawaybox}

\subsection{Optimal Post-Training Requires Balancing SFT and RL}

\begin{wrapfigure}{r}{0.58\textwidth}
    \centering
    \includegraphics[width=0.56\textwidth]{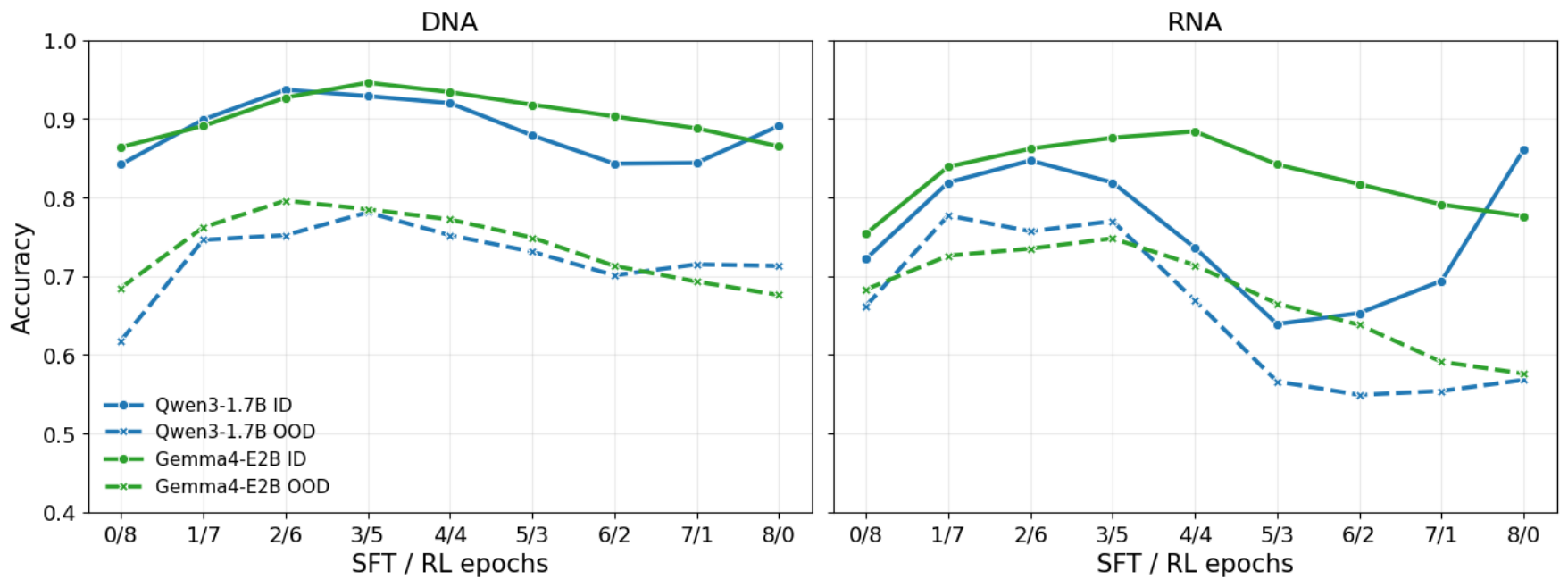}
    \caption{\textbf{Under a fixed post-training budget, a small amount of SFT followed by more RL gives the best ID-OOD trade-off.} Across DNA and RNA, 1--3 SFT epochs followed by larger RL budgets generally give the strongest OOD accuracy, while larger SFT allocations achieve better ID performance.}
    \label{fig:sft_rl_epoch_constraint}
\end{wrapfigure}


Finally, we study how to allocate post-training across supervised fine-tuning and reinforcement learning. In this experiment, we evaluate model families of the form {\color{mayablue}Q1-D/R-S}$_{\color{mayablue}s,1.5}${\color{mayablue}-R}$_{\color{mayablue}8-s,1.5}$ and {\color{green!60!black}G-D/R-S}$_{\color{green!60!black}s,1.5}${\color{green!60!black}-R}$_{\color{green!60!black}8-s,1.5}$, where the total post-training schedule is fixed at eight epoch-level passes, and only the split between SFT and RL is varied. This is not a strictly FLOP-matched comparison: an RL epoch is more expensive than an SFT epoch because GRPO uses multiple autoregressive rollouts, reward computation, and KL anchoring. We therefore interpret this setup as an epoch-budget allocation study that compares pure RL, pure SFT, and intermediate stage orderings under a common pass-count constraint, rather than as an exact compute-normalized optimum.

Figure~\ref{fig:sft_rl_epoch_constraint} shows that the best allocation is not at either extreme. In the DNA panel, the additional {\color{green!60!black}G-D} results closely mirror the {\color{mayablue}Q1-D} trend: ID accuracy is strongest after a few SFT epochs, peaking near \(0.95\) for {\color{green!60!black}G-D} and \(0.94\) for {\color{mayablue}Q1-D}, while OOD accuracy is maximized in the early mixed regime around \(0.78\). Pure RL underperforms these mixed schedules, especially OOD, indicating that reward optimization benefits from a supervised warm start. Pure SFT preserves relatively high ID accuracy, but its OOD performance is much weaker, falling to about \(0.71\) for {\color{mayablue}Q1-D} and \(0.68\) for {\color{green!60!black}G-D}. The {\color{green!60!black}G-D} curves also show stronger ID retention than {\color{mayablue}Q1-D} across several larger-SFT allocations, but this does not remove the OOD decline as SFT dominates the budget. The RNA panel shows the same qualitative trade-off more sharply: {\color{green!60!black}G-R} maintains higher ID accuracy across most allocations, whereas both models obtain their best OOD performance with only a small amount of SFT before RL. Overall, the strongest OOD results concentrate in the {\color{mayablue}Q1-R-S}$_{\color{mayablue}1,1.5}${\color{mayablue}-R}$_{\color{mayablue}7,1.5}$ to {\color{mayablue}Q1-R-S}$_{\color{mayablue}3,1.5}${\color{mayablue}-R}$_{\color{mayablue}5,1.5}$ and {\color{green!60!black}G-R-S}$_{\color{green!60!black}1,1.5}${\color{green!60!black}-R}$_{\color{green!60!black}7,1.5}$ to {\color{green!60!black}G-R-S}$_{\color{green!60!black}3,1.5}${\color{green!60!black}-R}$_{\color{green!60!black}5,1.5}$ range. This suggests that later post-training passes are better spent on RL once SFT has established task competence, although a compute-normalized study using estimated FLOPs would be needed to identify the exact optimal SFT--RL trade-off.

\begin{takeawaybox}
\textbf{Finding 6.} Under a fixed epoch-level post-training schedule, larger SFT allocations can preserve strong ID performance, but the best ID-OOD trade-off comes from a small amount of SFT followed by a larger RL allocation.
\end{takeawaybox}
\section{Discussion}

\begin{figure}[h]
    \centering
    \includegraphics[width=\textwidth]{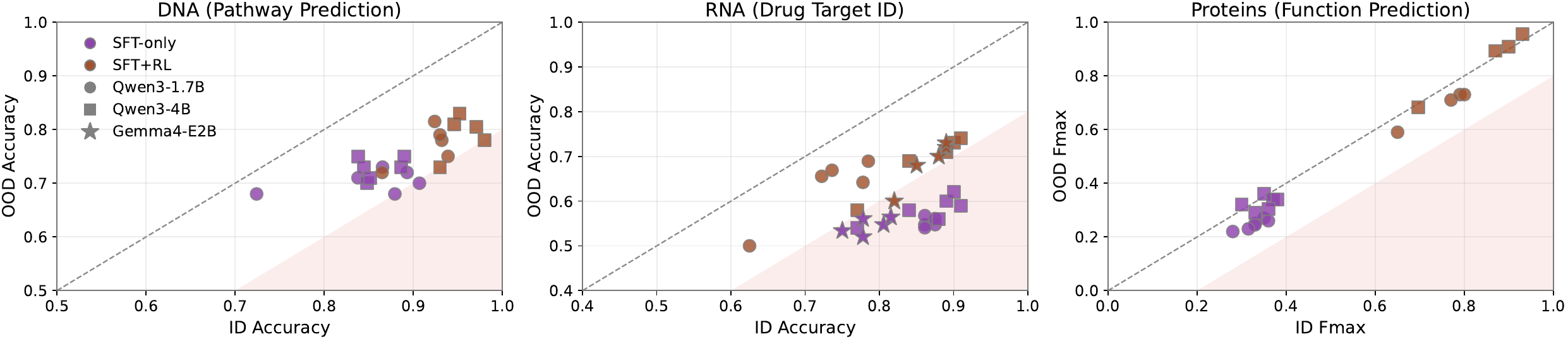}
    \caption{\textbf{RL shifts the ID-OOD frontier across modalities.}
    Each point is a trained checkpoint; color denotes training stage and marker shape denotes backbone. RL generally improves OOD performance at comparable ID performance across DNA, RNA, and protein tasks.}
    \label{fig:results_overview}
\end{figure}



Our results show that post-training stages play distinct roles in biological reasoning. CPT adapts models to biological language, SFT establishes task competence, and RL improves transfer beyond the training distribution. These stages should therefore not be treated as interchangeable sources of compute. Figure~\ref{fig:results_overview} summarizes this pattern across modalities, showing that RL shifts checkpoints toward stronger OOD performance at comparable ID performance. In practice, this suggests a simple recipe: use CPT to align models with biological language, use enough SFT to establish task competence, and allocate later post-training to RL when OOD robustness matters.


The SFT-induced trade-off also highlights why biology provides a demanding setting for studying post-training dynamics. In our RNA experiments, OOD accuracy drops by roughly 18 percentage points from its peak as SFT continues, following an approximately monotonic decline rather than a plateau. Biology exposes generalization failures that are often less apparent in conventional reasoning benchmarks. In mathematics and code, many OOD problems preserve the same underlying structure as the training examples, even when surface details change. In biology, unseen pathways, diseases, species, and perturbations often involve different mechanisms and biological processes. As a result, optimization that improves performance on the training distribution can simultaneously reduce the ability to transfer across biological systems.

\paragraph{Limitations and future work.}
Our study has several limitations. First, although we evaluate post-training across DNA, RNA, and protein reasoning tasks with biologically meaningful OOD splits, our conclusions rest on a limited set of tasks, benchmarks, and model families. It remains unclear how broadly these trends extend to other scientific reasoning settings and richer biological workflows. Second, our results suggest that RL depends on reward design, task structure, and the quality of the supervised starting point, but we have only begun to characterize these dependencies. Our fixed-schedule SFT-RL study is also not fully compute-normalized. Finally, our evaluation measures final-answer correctness rather than the validity of intermediate reasoning steps. We therefore cannot fully distinguish genuine biological reasoning from shortcut strategies that produce correct outputs. Future work should test these trade-offs on broader benchmarks and examine how reward design, compute-normalized stage allocation, and adaptation capacity shape ID-OOD robustness in scientific reasoning models.

More broadly, our results suggest that progress in scientific reasoning will depend not only on larger models or more post-training compute, but on understanding how different stages shape generalization. In biology, the strongest models are not those that optimize longest on a fixed distribution, but those that preserve the ability to transfer across biological systems. Understanding and controlling these training dynamics may therefore be as important as scaling model size itself.

\subsection*{Acknowledgements}

{\small L.F. is supported by the Kempner Graduate Fellowship at Harvard University. H.Z. and S.K. acknowledge the Chan Zuckerberg Initiative Foundation for establishing the Kempner Institute for the Study of Natural and Artificial Intelligence. 
M.M.L. and M.Z. gratefully acknowledge the support, in part, by NSF CAREER Award 2339524, ARPA-H Biomedical Data Fabric (BDF) Toolbox Program, Amazon Faculty Research, Google Research Scholar Program, AstraZeneca Research, GlaxoSmithKline Award, Roche Alliance with Distinguished Scientists (ROADS) Program, Sanofi iDEA-iTECH Award, Boehringer Ingelheim Award, Merck Award, Optum AI Research Collaboration Award, Pfizer Research, Gates Foundation (INV-079038), Chan Zuckerberg Initiative, Collaborative Center for XDP at Massachusetts General Hospital, John and Virginia Kaneb Fellowship at Harvard Medical School, Biswas Computational Biology Initiative in partnership with the Milken Institute, Harvard Medical School Dean's Innovation Fund for the Use of Artificial Intelligence, and the Kempner Institute for the Study of Natural and Artificial Intelligence at Harvard University. 
Any opinions, findings, conclusions or recommendations expressed in this material are those of the authors and do not necessarily reflect the views of the funders.

Authors affiliated with Google DeepMind and Google Research (Eric Wang, Shekoofeh Azizi, Bryan Perozzi) participated in this work in an advisory capacity only.
}

\bibliographystyle{unsrtnat}
\bibliography{main}


\clearpage
\appendix
\etocdepthtag.toc{appendix}

\etocsettagdepth{main}{none}
\etocsettagdepth{appendix}{subsection}

\renewcommand{\contentsname}{Appendices}
\tableofcontents
\clearpage

\section{Additional Details on Biological Reasoning Tasks}
\label{app:datasets}

\subsection{Pathway Prediction Dataset and Construction}

Our pathway prediction benchmark follows the KEGG-derived reasoning dataset introduced in BIOREASON~\citep{fallahpour2025bioreason}. The source data begins from KEGG Network Variants and associated disease-pathway annotations~\citep{kanehisa2000kegg,kanehisa2019keggnetwork}, which BIOREASON then augments with linked variant metadata from ClinVar, dbSNP, OMIM, and COSMIC~\citep{landrum2020clinvar,sherry2001dbsnp,amberger2015omim,tate2019cosmic}. The resulting benchmark contains 1,449 examples spanning 298 pathway networks and 37 unique diseases. In the original curation, the key design goal was not simply to label variants, but to preserve the mechanistic chain from mutation to pathway perturbation to phenotype, so that each example could support explicit multi-step biological reasoning rather than endpoint classification alone.

Technically, each KEGG pathway is represented as a structured molecular interaction network using a standardized symbolic notation that encodes activation, inhibition, complex formation, and transcriptional regulation~\citep{kanehisa2000kegg,kanehisa2019keggnetwork}. These pathway graphs are then linked to specific variants through a semi-automated mapping procedure designed to preserve the relationship between genomic loci and pathway entities. For each mapped example, the dataset stores paired reference and variant DNA sequences with precise alignment coordinates; in BIOREASON, these sequences average roughly 4,000 base pairs, with most mutations differing from the reference by only 1--3 nucleotides. The final supervised example consists of variant details, a network definition, and gene-level context on the input side, together with a concise mechanism-to-disease answer and a full reasoning trace on the output side.

A distinctive part of the construction is the generation of causal reasoning paths. In BIOREASON, these traces were produced using Claude 3.7 Sonnet and grounded with contextual disease information from the KEGG disease database~\citep{kanehisa2000kegg,kanehisa2019keggnetwork}, then packaged into standardized question-answer pairs for training and evaluation~\citep{fallahpour2025bioreason}. The reasoning traces have mean length 303.8 words and are intended to make the latent biological mechanism explicit: they verbalize how the mutation perturbs the affected gene, how that perturbation propagates through intermediate pathway interactions, and why the resulting network state is associated with the target disease. In our work, we inherit this benchmark structure but evaluate generalization by splitting at the level of pathway networks, so that out-of-domain examples require transfer to previously unseen molecular systems rather than new variants within familiar ones.

\subsection{Target Identification Dataset and Simplifications Relative to MEDEA}

Our target identification benchmark is adapted from the cell type specific target nomination task introduced in MEDEA~\citep{sui2026medea}, but we convert it from an agentic workflow into a fixed-input reasoning problem. In the original MEDEA setup, the task is defined over five diseases---rheumatoid arthritis, type 1 diabetes mellitus, Sjogren's syndrome, hepatoblastoma, and follicular lymphoma---and 29 cell types, with each analysis asking the model to select the best therapeutic target from a set of five candidate genes for a specified disease--cell-type context. The full benchmark contains 2,400 analyses in total, generated from disease atlases and target-disease resources to test whether models can identify therapeutically plausible targets at cell-type resolution rather than from bulk tissue averages.

The dataset construction in MEDEA proceeds in three stages. First, each disease atlas is processed from CELLxGENE~\citep{cziscience2025cellxgene} using a standard single-cell pipeline, followed by one-vs-all differential expression analysis to identify disease-specific marker genes for each cell type and disease combination. Second, disease-associated genes are collected from Open Targets~\citep{ochoa2023opentargets}, keeping genes with nonzero genetic evidence or ChEMBL evidence~\citep{gaulton2012chembl}. Third, ground-truth cell type specific disease targets are defined as genes satisfying both criteria: they are differentially expressed in the relevant disease--cell-type context and supported by disease-target evidence. For each context, MEDEA then forms five-gene candidate sets by sampling one positive target and four negatives, and uses prompt paraphrasing plus multiple random seeds to generate the final benchmark.

Our version keeps the same core supervision signal but simplifies the task substantially relative to MEDEA~\citep{sui2026medea}. Rather than asking an agent to construct a research plan, invoke tools, retrieve literature, and reconcile evidence across multiple modules, we provide the model with the disease, cell type, candidate genes, and transcriptomic evidence directly. Concretely, instead of tool-based retrieval over single-cell atlases and other external resources, we supply aligned TranscriptFormer embeddings~\citep{pearce2025cross} for the five candidate genes in both normal and disease states for the relevant context. This removes the planning, execution, and literature-reasoning burden while preserving the central inferential challenge: the model must still identify which candidate is most supported in the specified disease and cell type, now using a fixed multimodal input rather than an open-ended agentic pipeline. Consistent with the main text, we further use four of the five diseases as the in-domain pool for train/ID-test splitting and reserve hepatoblastoma as the OOD test disease.

\subsection{Protein Function Prediction Dataset and Simplifications Relative to BioReason-Pro}

Our protein function prediction benchmark is adapted from the curated UniProt-based dataset introduced in BioReason-Pro~\citep{fallahpour2026bioreason}. The original corpus is designed around experimentally supported protein function annotation rather than generic sequence-level pretraining, and integrates multiple biological modalities into a single example. Starting from UniProt and the GOA database~\citep{uniprot2023uniprot,huntley2015goa}, BioReason-Pro retains only proteins with experimental or curated GO evidence codes, standardizes annotations to the January 2023 Gene Ontology~\citep{geneontology2023}, and propagates terms upward through the ontology hierarchy to preserve hierarchical completeness. The resulting dataset contains 133,492 proteins spanning 3,135 organisms, with each protein linked not only to its amino acid sequence, but also to organism metadata, subcellular localization, InterPro domain annotations~\citep{paysan2023interpro}, structural information, and protein--protein interaction context~\citep{szklarczyk2025string}.

At the instance level, BioReason-Pro constructs a compact multimodal context for each protein by combining InterPro domains with residue ranges~\citep{paysan2023interpro}, the UniProt protein description~\citep{uniprot2023uniprot}, organism, subcellular localization, STRING interaction partners~\citep{szklarczyk2025string}, and GO leaf terms across molecular function, biological process, and cellular component~\citep{geneontology2023}. These contexts are then used to generate synthetic step-by-step reasoning traces, which end in a structured final answer containing a concise function summary, the relevant InterPro domains, GO terms, and an interaction hypothesis. Evaluation follows the CAFA temporal holdout protocol~\citep{radivojac2013large,zhou2019cafa}: proteins annotated before November 2022 are used for training and validation, while test proteins are selected from those that gained new experimental annotations between March 2023 and February 2024 and lacked annotations in the target aspect beforehand. The final temporal holdout test set contains 8,630 proteins and 230,824 propagated GO annotations.

Our version keeps the same overall prediction task and temporal evaluation logic, but simplifies the original BioReason-Pro setup to match the common multimodal format used throughout this paper~\citep{fallahpour2026bioreason}. In BioReason-Pro itself, the model consumes residue-level ESM3 embeddings~\citep{hayes2025simulating}, organism and textual biological context, GO-GPT predictions, and an additional GO graph encoder that injects explicit ontology structure into the language model~\citep{geneontology2023}. In our benchmark, we remove this GO graph input and treat the task as reasoning from protein representations plus textual biological metadata alone. Concretely, the model receives protein embeddings together with text context such as organism, InterPro domains, and protein--protein interactions, and must infer function without direct access to ontology graph embeddings. This simplification preserves the core challenge of integrating sequence-derived and symbolic evidence for protein function prediction, while making the protein task architecturally comparable to the DNA and RNA settings studied in the main text.

\subsection{Example Prompts and Inputs for Each Task}

\paragraph{Pathway prediction.} A representative input consists of two versions of the same DNA sequence region, with and without the mutation, including DNA-specific start and padding tokens, followed by a pathway network definition and gene annotations. The model is then asked to infer the biological or disease effect associated with the allele. For example:

\begin{quote}
\texttt{<|dna\_pad|>...<|dna\_pad|>}

\textbf{Question:} Network Definition of the pathway: \texttt{SOD1* -| BIP -| ERN1 -> XBP1}; Genes in the pathway: \textit{SOD1}; superoxide dismutase 1 | \textit{HSPA5}; heat shock protein family A (Hsp70) member 5 | \textit{ERN1}; endoplasmic reticulum to nucleus signaling 1 | \textit{XBP1}; X-box binding protein 1. Given this context, what is the biological effect of this \textit{SOD1} allele, specifically what disease does this contribute to?

\textbf{Answer:} amyotrophic lateral sclerosis
\end{quote}

\paragraph{Drug target identification.} A representative input consists of prepended RNA padding tokens followed by a disease- and cell-type-specific target selection question over candidate genes. For example:

\begin{quote}
\texttt{<|rna\_pad|>...<|rna\_pad|>}

\textbf{Question:} Among the genes \textit{PDRG1}, \textit{PIK3CG}, \textit{TRIM23}, \textit{SUCO}, and \textit{GCKR}, which one exhibits the highest T cell-specific expression relevant for targeted intervention in follicular lymphoma?

\textbf{Answer:} \textit{PIK3CG}
\end{quote}

The five projected TranscriptFormer representations are inserted before the text prompt in the same order as the five candidate gene names listed in the question; this fixed ordering defines the alignment between continuous RNA embeddings and discrete gene symbols.

\paragraph{Protein function prediction.} A representative input consists of protein-specific padding tokens, followed by organism metadata, InterPro domain annotations, optional protein--protein interaction context, and initial GO term speculations. For example:

\begin{quote}
\texttt{<|im\_start|>user}

\textbf{Protein:} \texttt{<|protein\_pad|><|protein\_pad|>\ldots<|protein\_pad|>}

You are a scientific assistant specialized in protein function prediction. Given a protein sequence, organism information, InterPro domain annotations, protein--protein interaction partners, and initial GO term speculations, reason about the function of the protein and summarize in UniProt format.

\textbf{Organism:} \textit{Homo sapiens}

\textbf{InterPro annotations:}
\begin{itemize}
    \item \texttt{IPR000795}: Translational tr-type GTP-binding domain (domain) [5--217]
    \item \texttt{IPR027417}: P-loop containing nucleoside triphosphate hydrolase (homologous superfamily) [5--239]
    \item \texttt{IPR004161}: Translation elongation factor EFTu-like, domain 2 (domain) [238--301]
    \item \texttt{IPR049393}: Selenocysteine-specific elongation factor, 3rd domain (domain) [317--448]
    \item \texttt{IPR049394}: Selenocysteine-specific elongation factor, C-terminal RIFT domain (domain) [465--590]
    \item \texttt{IPR050055}: Elongation factor Tu GTPase family (family) [5--445]
\end{itemize}

\textbf{Protein--protein interaction partners:} None provided.

\textbf{Initial GO term speculations:}

Molecular Function (MF): \texttt{GO:0003924} GTPase activity, \texttt{GO:0005525} GTP binding, \texttt{GO:0003746} translation elongation factor activity

Biological Process (BP): \texttt{GO:0006415} translational elongation, \texttt{GO:0006412} translation

Cellular Component (CC): \texttt{GO:0005829} cytosol

Reason about the function of the protein and focus on Molecular Function, Biological Process, and Cellular Component.

\texttt{<|im\_end|>}

\texttt{<|im\_start|>assistant}

\textbf{Answer:} This protein is a selenocysteine-specific translation elongation factor that uses GTP binding and hydrolysis to deliver selenocysteinyl-tRNA\textsuperscript{Sec} during translational elongation. A plausible functional interaction partner is SECIS-binding protein 2 (SBP2).
\end{quote}

\section{Experimental Setup}
\label{app:setup}

\subsection{Base Models, Tokenization, and Input Representations}
\label{app:setup-models}

\paragraph{DNA experiments.}
The text backbones for DNA continued pre-training (CPT) and the downstream SFT/RL stages are Qwen3-1.7B and Qwen3-4B, loaded with their native tokenizers~\citep{yang2025qwen3}. For CPT we train the language model alone on tokenized biological free-text (no DNA encoder is attached). For the post-CPT SFT and RL stages we couple the text backbone to a frozen Evo2-1B encoder via a trainable linear projection~\citep{brixi2026genome}; the DNA hidden state is prepended to the text embeddings. DNA sequences are clipped to a maximum length of $2048$ nucleotides, with $1024$ nucleotides retained on each flank around the variant locus.

\paragraph{RNA experiments.}
The RNA experiments follow the same overall setup as the DNA experiments, replacing the sequence encoder and biological input modality while keeping the same text backbones and tokenizer choices. The text backbones for RNA CPT, SFT, and RL are Qwen3-1.7B and Qwen3-4B, loaded with their native tokenizers~\citep{yang2025qwen3}. As in the DNA setting, CPT is performed on tokenized biological free-text using the language model alone, without attaching the RNA encoder.

For the downstream SFT and RL stages, we couple the text backbone to a frozen TranscriptFormer encoder through a trainable linear projection~\citep{pearce2025cross}. Each target-identification example contains a natural-language disease and cell-type context, a five-gene candidate set, and TranscriptFormer representations for the candidate genes in normal and disease states~\citep{pearce2025cross,sui2026medea}. The projected RNA hidden states are prepended to the text-token embeddings before the prompt tokens, so that the language model conditions jointly on transcriptomic representations and the textual task description.

\paragraph{Protein experiments.}
The text backbones for the protein experiments are Qwen3-1.7B and Qwen3-4B-Thinking, loaded with their native BPE tokenizer~\citep{yang2025qwen3}. The padding token is aliased to the end-of-sequence token. Both the SFT and RL prompts concatenate (i)~a system instruction describing the task and available biological context, (ii)~a header containing the protein name, organism, and amino-acid sequence, and (iii)~a user instruction asking the model to emit GO identifiers across the molecular function, biological process, and cellular component aspects~\citep{geneontology2023}.

For SFT and GRPO we use the same BioReason-Pro-style protein-conditioned interface, except that we omit the GO-graph encoder~\citep{fallahpour2026bioreason}. The text backbone is paired with a frozen ESM-3 small protein encoder~\citep{hayes2025simulating}. Per-residue embeddings are extracted from layer 37 of ESM-3, projected through a trainable linear layer into the text-embedding space, and inserted at the protein placeholder positions before the text tokens~\citep{hayes2025simulating}. The protein encoder is kept frozen; the protein projection layer and the LoRA adapter on the text model receive gradients~\citep{hu2022lora}.

\subsection{Continued Pre-training (Mid-training) Setup}
\label{app:setup-cpt}

We mid-train the two Qwen3 backbones on the biology subset of FineFineWeb~\citep{map2024finefineweb}. We use the first
200,000 documents for training and hold out the next 5,000 as a fixed evaluation set, yielding a
200K/5K train/eval split.

We do not perform additional benchmark-specific deduplication or decontamination of the FineWeb
biology subset beyond the filtering already implicit in the source corpus~\citep{penedo2024fineweb,map2024finefineweb}. Our goal in CPT is to model
a realistic domain-adaptation setting starting from publicly available pretrained LLMs, whose original
pretraining corpora are not fully auditable and may already contain task-relevant biological text~\citep{gururangan2020don,que2024d,gupta2023continual}. We
therefore treat CPT as exposure to broad biological language rather than as a controlled from-scratch
pretraining intervention. Importantly, the CPT corpus does not include our supervised reasoning
traces, RL prompts, or any constructed train/test examples from the downstream benchmarks. The
main leakage-sensitive comparisons in the paper are stage-wise and relative: all CPT and non-CPT
models use the same downstream splits, and OOD evaluation is defined by held-out pathways,
diseases, or species. We therefore interpret CPT results as measuring the effect of additional broad
biological language adaptation under realistic pretrained-model conditions, not as evidence of strict
benchmark decontamination.

Training uses the standard causal-LM next-token prediction loss. Inputs are tokenized with a maximum length of $1024$ tokens. We optimize with AdamW under a cosine learning-rate schedule with $3\%$ linear warm-up, weight decay $0.01$, gradient clipping $1.0$, and bf16 mixed precision~\citep{loshchilov2019decoupled}. Each run trains for one epoch with a per-device batch size of~$1$. The CPT hyperparameter sweep varies the learning rate over ${1\!\times\!10^{-5}, 3\!\times\!10^{-4}}$ and the gradient-accumulation steps over ${64,128}$ for both backbones. We select the best checkpoint by validation loss.

\subsection{Supervised Fine-Tuning Setup}
\label{app:setup-sft}

\paragraph{DNA SFT.}
DNA SFT couples a frozen Evo2-1B encoder with either Qwen3-1.7B or Qwen3-4B (including their post-CPT variants; see~\S\ref{app:setup-cpt})~\citep{brixi2026genome,yang2025qwen3}. We use DeepSpeed Stage~2 on a single GPU with bf16 precision, batch size~$1$, gradient accumulation~$8$, AdamW with $\eta=5\times10^{-5}$ and weight decay~$0.01$, and the same warm-up-to-cosine schedule as the protein SFT ($5\%$ warm-up, decay floor $0.1\eta_{\max}$)~\citep{rajbhandari2020zero,loshchilov2019decoupled}. The maximum DNA sequence length is $2048$ nucleotides and the maximum text length is $1024$ tokens. The epoch sweep ranges over $\{1,2,4,8,16,32\}$ for each of $\{$Qwen3-1.7B, Qwen3-4B, CPT-Qwen3-1.7B, CPT-Qwen3-4B$\}$, with the best CPT learning rate selected per backbone from~\S\ref{app:setup-cpt}. Random seed is~$23$.

\paragraph{RNA SFT.}
RNA SFT follows the same training recipe as DNA SFT, replacing the Evo2 DNA encoder with the frozen TranscriptFormer encoder and using the target-identification examples described in~\S\ref{app:setup-models}~\citep{pearce2025cross,sui2026medea}. Each example provides a disease, cell type, five candidate genes, and TranscriptFormer embeddings for the corresponding normal and disease states~\citep{pearce2025cross}. The projected TranscriptFormer hidden states are prepended to the text-token embeddings, and the model is trained to generate the correct target gene.

We use DeepSpeed Stage~2 on a single GPU with bf16 precision, batch size~$1$, gradient accumulation~$8$, AdamW with $\eta=5 \times 10^{-5}$ and weight decay~$0.01$, and $5\%$ linear warm-up followed by cosine decay to $0.1\eta_{\max}$~\citep{rajbhandari2020zero,loshchilov2019decoupled}. The maximum text length is $1024$ tokens. The main epoch sweep ranges over $\{1,2,4,8,16,32\}$ for each of Qwen3-1.7B and Qwen3-4B, using the full RNA training set~\citep{yang2025qwen3}. The same sweep is repeated for the Gemma 4 E2B RNA backbone in the backbone ablation~\citep{deepmind2026gemma4}. Random seed is~$23$.

\paragraph{Protein SFT.}
SFT pairs the frozen ESM-3 small encoder with the Qwen3-1.7B or Qwen3-4B-Thinking text backbone (see~\S\ref{app:setup-models})~\citep{hayes2025simulating,yang2025qwen3}.

We use single-GPU training with bf16 mixed precision, batch size~$1$, and gradient accumulation~$16$ (effective batch size~$16$). Optimization uses AdamW with $\eta_{\max}=1\!\times\!10^{-4}$, weight decay $0.01$, $5\%$ linear warm-up followed by cosine decay to $0.1\eta_{\max}$~\citep{loshchilov2019decoupled}. The maximum text sequence length is $10{,}000$ tokens and the maximum protein length is $2000$ residues. Flash attention is enabled where supported, along with gradient checkpointing~\citep{dao2022flashattention}. Validation is run at the end of each epoch on a $10\%$ held-out split, and we keep the single best checkpoint by validation loss. Two sweeps are run: a \emph{data fraction sweep} at $1$ epoch over $\{20, 40, 60, 80, 100\}\%$ of the training data, and an \emph{epoch sweep} at $20\%$ data over $\{1,2,4,8,16,32\}$ epochs. Random seed is fixed to $23$.

\subsection{Reinforcement-Learning (GRPO) Setup}
\label{app:setup-rl}

\paragraph{DNA RL.}
The DNA GRPO runs use a multimodal architecture wrapping Qwen3-1.7B/4B (or their CPT variants) with a frozen Evo2-1B encoder~\citep{guo2025deepseek,yang2025qwen3,brixi2026genome}. The text model receives a fresh LoRA adapter attached on top of the SFT-merged checkpoint (see~\S\ref{app:setup-lora}); the DNA encoder is frozen and the DNA projection is trainable~\citep{hu2022lora}. We use DeepSpeed Stage~2 with bf16 precision on a single GPU~\citep{rajbhandari2020zero}. Generation uses $8$ rollouts per prompt with max completion length $800$, $T=1$, top-$p=0.95$, and top-$k=20$. Optimization runs at $\eta=1\times10^{-5}$ with a cosine schedule, $3\%$ warm-up, per-device batch size $4$, gradient accumulation $8$, gradient checkpointing, and $\beta=10^{-4}$ for the KL anchor~\citep{ouyang2022training}. The reward combines format-adherence and correctness components following \citep{fallahpour2025bioreason}. The number of GRPO epochs is swept over $\{1,2,4,8,16\}$; each RL run selects the best SFT checkpoint by validation accuracy, merges the SFT LoRA into the base weights, and then attaches a fresh RL adapter.

\paragraph{RNA RL.}
RNA GRPO follows the same multimodal RL setup as DNA GRPO, replacing the Evo2-1B encoder with the frozen TranscriptFormer encoder and using the target-identification reward~\citep{guo2025deepseek,pearce2025cross,sui2026medea}. The policy wraps Qwen3-1.7B, Qwen3-4B, or the Gemma 4 E2B RNA backbone with the frozen TranscriptFormer encoder and a trainable projection layer~\citep{yang2025qwen3,deepmind2026gemma4,pearce2025cross}. As in the DNA setting, the SFT LoRA is first merged into the text backbone, after which a fresh RL LoRA adapter is attached for GRPO~\citep{hu2022lora}. The TranscriptFormer encoder remains frozen throughout RL, while the projection layer and RL adapter are trainable.

For each prompt, the model receives the disease, cell type, five candidate genes, and projected TranscriptFormer representations for the corresponding normal and disease states~\citep{pearce2025cross,sui2026medea}. The reward contains a format-adherence term and a correctness term based on whether the final answer matches the held-out target gene. For the main Qwen RNA sweeps, GRPO starts from the strongest SFT checkpoints identified in the SFT sweep: Qwen3-1.7B-R-SFT4,1 and Qwen3-4B-R-SFT8,1. The number of GRPO epochs is swept over $\{1,2,4,8,16\}$, yielding model families of the form Qwen3-1.7B-R-SFT4,1-RL$\{1,2,4,8,16\}$,1 and Qwen3-4B-R-SFT8,1-RL$\{1,2,4,8,16\}$,1. For the backbone ablation, the same procedure is applied to Gemma 4 E2B-R-SFT4,1-RL$\{1,2,4,8,16\}$,1.

Unless otherwise stated, RNA GRPO uses the same optimization and generation hyperparameters as DNA GRPO: DeepSpeed Stage~2, bf16 precision, $8$ rollouts per prompt, max completion length $800$, $T=1$, top-$p=0.95$, top-$k=20$, AdamW with $\eta=1\times10^{-5}$, cosine decay with $3\%$ warm-up, per-device batch size $4$, gradient accumulation $8$, gradient checkpointing, and KL coefficient $\beta=10^{-4}$~\citep{rajbhandari2020zero,loshchilov2019decoupled,guo2025deepseek}. The random seed is fixed to $23$.

For evaluation, the generated final answer is normalized by case, whitespace, punctuation, and common gene-symbol formatting variants, and is counted correct only if it exactly matches the held-out target gene symbol. Following \citep{fallahpour2025bioreason}, the model gets additional rewards for following the expected output structure, i.e. a reasoning trace followed by a one-gene answer, and for staying below the token limit of 1024 tokens (conciseness reward).

\paragraph{Protein RL.}
We apply Group Relative Policy Optimization (GRPO) to the same protein-conditioned Qwen3-1.7B/Qwen3-4B-Thinking policy used in SFT~\citep{guo2025deepseek,yang2025qwen3}. For each prompt, frozen ESM-3 residue embeddings are projected into the Qwen embedding space and inserted at the protein placeholder positions; unlike BioReason-Pro, no GO-graph embeddings are included~\citep{hayes2025simulating,fallahpour2026bioreason}. All GRPO runs warm-start from the SFT-trained LoRA adapter and protein projection extracted from the matching SFT checkpoint; the no-warm-start ablation trains from scratch~\citep{hu2022lora}. A frozen copy of this SFT-initialized protein-conditioned policy serves as the reference policy for the KL anchor~\citep{ouyang2022training}. Gradient checkpointing is enabled on the policy to control activation memory.

For each minibatch of $b$ prompts, we draw $g$ rollout completions per prompt with sampling temperature $T$ and top-$p$, and compute the propagated GO-F1 reward
\[
r_{i,k}
=
\mathrm{F1}\!\left(
\mathcal{G}(\hat{y}_{i,k}),
\mathcal{G}(y_i^{*})
\right),
\]
where $\mathcal{G}(\cdot)$ denotes the set of ontology-propagated GO terms extracted from a completion or gold answer~\citep{geneontology2023}. We form group-centered, batch-standard-deviation-normalized advantages
\[
\hat{A}_{i,k}
=
\frac{r_{i,k}-\bar{r}_i}{\sigma_r+\epsilon_{\mathrm{adv}}},
\qquad
\bar{r}_i
=
\frac{1}{g}\sum_{k=1}^{g} r_{i,k}.
\]
The objective is the per-token clipped surrogate of GRPO with an unbiased $k_3$ KL anchor against the frozen reference policy~\citep{schulman2017proximal,guo2025deepseek}:
\[
\mathcal{L}_{\mathrm{GRPO}}
=
-\mathbb{E}_{i,k,t}
\left[
\min\!\left(
\rho_{i,k,t}\,\hat{A}_{i,k},
\operatorname{clip}\!\left(
\rho_{i,k,t},
1-\epsilon_{\mathrm{lo}},
1+\epsilon_{\mathrm{hi}}
\right)\hat{A}_{i,k}
\right)
-
\beta\,\mathrm{KL}_{k_3,t}\!\left(\pi_\theta \,\|\, \pi_{\mathrm{ref}}\right)
\right],
\]
where
\[
\rho_{i,k,t}
=
\exp\!\left(
\log \pi_{\theta}\!\left(y_{i,k,t}\mid x_i,y_{i,k,<t}\right)
-
\log \pi_{\theta_{\mathrm{old}}}\!\left(y_{i,k,t}\mid x_i,y_{i,k,<t}\right)
\right).
\]
The old-policy log probabilities $\log \pi_{\theta_{\mathrm{old}}}$ are computed under the pre-update rollout policy and cached or recomputed without gradient flow before the policy update; they are detached only for importance weighting and are not set equal to the current-policy numerator. The per-token $k_3$ estimator is
\[
\mathrm{KL}_{k_3,t}
=
\exp(u_t)-u_t-1,
\qquad
u_t
=
\left(
\log \pi_{\mathrm{ref}}
-
\log \pi_{\theta}
\right)_t,
\]
with $u_t$ clamped to $[-20,20]$ for numerical safety.

We sweep over $\beta\in{10^{-4},,10^{-3}}$. Other GRPO settings are: $\epsilon_{\mathrm{lo}}=0.20$, $\epsilon_{\mathrm{hi}}=0.28$, $\epsilon_{\mathrm{adv}}=10^{-6}$, batch size $b=1$, group size $g=2$, max new tokens $256$, $T=1.0$, top-$p=1.0$. The optimizer is AdamW with $\eta=3\times10^{-5}$, weight decay $0.01$, gradient clipping $1.0$, and bf16 precision~\citep{loshchilov2019decoupled}. The training data is capped to match the SFT data fractions $\{20,40,60,80,100\}\%$ of $20{,}000$ examples. We include InterPro features in the prompt but exclude protein--protein interactions~\citep{paysan2023interpro,szklarczyk2025string}. The reward is computed against ontology-propagated leaf GO terms; examples without resolvable gold terms are dropped~\citep{geneontology2023}. Both the GRPO reward and the evaluation metric are propagated unweighted GO-F1~\citep{fallahpour2026bioreason,geneontology2023}.

\paragraph{RL evaluation.}
For each trained run we evaluate on both the ID and OOD test splits. We select the checkpoint with the highest centred 50-step rolling mean of the training reward, as several runs at $\beta=10^{-4}$ reach peak reward mid-training before drifting. Generation uses sampling at $T=0.7$ with max new tokens $512$; greedy decoding is avoided because the deterministic path on Qwen3-4B-Thinking tends to remain in the reasoning block without emitting a final answer.

\subsection{LoRA Configurations and Trainable-Parameter Choices}
\label{app:setup-lora}

All LoRA adapters target the seven attention and MLP projection matrices per transformer block: {q, k, v, o, gate, up, down} projections, with Gaussian initialization and no bias~\citep{hu2022lora}.

\paragraph{DNA LoRA.}
Both the DNA SFT and RL stages use $r=32$, $\alpha=64$, dropout~$0.05$ on the same seven target modules. The Evo2 encoder is frozen; the DNA projection is trainable~\citep{brixi2026genome}. For RL, when the SFT and RL adapter ranks differ, we merge the SFT LoRA into the base weights and attach a fresh RL adapter at the requested rank~\citep{hu2022lora}.

\paragraph{RNA LoRA.}
The RNA SFT stage uses $r=64$, $\alpha=64$, dropout~$0.05$, while the RL stage uses $r=16$, $\alpha=64$, dropout~$0.05$ on the same seven target modules. The transcriptformer encoder is frozen; the RNA projection is trainable~\citep{pearce2025cross}. For RL, when the SFT and RL adapter ranks differ, we merge the SFT LoRA into the base weights and attach a fresh RL adapter at the requested rank~\citep{hu2022lora}.

\paragraph{Protein SFT LoRA.}
Rank $r=128$, scaling $\alpha=256$, dropout~$0$. ESM-3 and the GO graph components are frozen; only the LoRA adapter on the text model and the protein-to-text projection layer receive gradients~\citep{hayes2025simulating,geneontology2023,hu2022lora}.

\paragraph{Protein RL warm-start.}
The SFT LoRA adapter and protein projection are extracted from the SFT checkpoint and reattached to the protein-conditioned Qwen policy used for GRPO~\citep{hu2022lora,guo2025deepseek}. ESM-3 remains frozen~\citep{hayes2025simulating}. During GRPO, the LoRA adapter and protein projection are trainable, while the underlying Qwen3-4B-Thinking weights are frozen~\citep{yang2025qwen3}. The no-warm-start ablation trains the full ~4 B-parameter model directly without a LoRA adapter.

\subsection{Hyperparameters, Context Windows, Sequence Lengths, and Optimization}
\label{app:setup-hp}

Table~\ref{tab:hp-summary} summarizes the configuration used to produce every reported result. All training uses AdamW with $\beta_1=0.9$, $\beta_2=0.999$, $\varepsilon=10^{-8}$. Random seed is~$23$ for protein and RNA experiments and~$42$ for DNA RL. All runs use bf16 mixed precision with flash attention where supported.

\begin{table}[h]
\centering
\small
\caption{Optimization and context-window settings per training stage. ``warm$\to$cos'' denotes linear warm-up followed by cosine decay to $\eta_{\min}$.}
\label{tab:hp-summary}
\resizebox{\textwidth}{!}{%
\begin{tabular}{lcccccc}
\toprule
Stage & Backbone(s) & $\eta_{\max}$ & Schedule & Warm-up & Batch (eff.) & Seq.\ len. \\
\midrule
DNA/RNA CPT & Qwen3-1.7B/4B & $\{10^{-5},\,3\!\times\!10^{-4}\}$ & cosine & $3\%$ & $1\!\times\!\{64,128\}$ & $1024$ \\
Protein SFT & Qwen3-1.7B/4B-Thinking & $10^{-4}$ & warm$\to$cos & $5\%$ & $1\!\times\!16$ & $10\text{K}\,/\,2\text{K}$\textsuperscript{$\dagger$} \\
DNA SFT & Qwen3-1.7B/4B (+CPT) & $5\!\times\!10^{-5}$ & warm$\to$cos & $5\%$ & $1\!\times\!8$ & $1024\,/\,2048$\textsuperscript{$\ddagger$} \\
RNA SFT & Qwen3-1.7B/4B, Gemma 4 E2B & $5\!\times\!10^{-5}$ & warm$\to$cos & $5\%$ & $1\!\times\!8$ & $1024$\textsuperscript{$\S$} \\
Protein GRPO & Qwen3-1.7B/4B-Thinking & $3\!\times\!10^{-5}$ & constant & --- & $1\!\times\!1$, $g\!=\!2$ & gen.\ $256$ \\
DNA GRPO & Qwen3-1.7B/4B (+CPT) & $10^{-5}$ & cosine & $3\%$ & $4\!\times\!8$, $g\!=\!8$ & gen.\ $800$ \\
RNA GRPO & Qwen3-1.7B/4B, Gemma 4 E2B & $10^{-5}$ & cosine & $3\%$ & $4\!\times\!8$, $g\!=\!8$ & gen.\ $1024$ \\
\bottomrule
\end{tabular}}\\[2pt]
{\small\textsuperscript{$\dagger$}\,Text\,/\,protein-residue length caps.\quad
\textsuperscript{$\ddagger$}\,Text\,/\,DNA length caps (truncated symmetrically around the variant locus).\quad
\textsuperscript{$\S$}\,Text length cap; TranscriptFormer embeddings are prepended separately.}
\end{table}

Weight decay is $0.01$ on every stage; gradient clipping is $1.0$. The protein GRPO step uses a constant learning rate; we rely on KL regularization and early-stopping checkpoint selection to control late-stage drift.

\paragraph{Sweep grids.}
\begin{itemize}
\item \textbf{DNA/ RNA CPT}: learning rate $\in\{10^{-5},\,3\!\times\!10^{-4}\}$, gradient accumulation $\in\{64,128\}$, $1$ epoch ($8$ runs total).
\item \textbf{Protein SFT data sweep}: data fraction $\in\{20,40,60,80,100\}\%$ at $1$ epoch.
\item \textbf{Protein RL data sweep}: matched data fractions at $1$ epoch with $\beta=10^{-4}$ warm-start; the $\beta=10^{-3}$ ablation re-runs the same grid.
\item \textbf{Protein RL epoch sweep}: epochs $\in\{1,2,4,8\}$ at $20\%$ data.
\item \textbf{DNA/ RNA SFT/RL epoch sweep}: epochs $\in\{1,2,4,8,16,32\}$ for SFT (capped at $8$ for RL), across Qwen3-1.7B, Qwen3-4B, and their post-CPT variants.
\end{itemize}

\subsection{Compute Resources and Training Budget}
\label{app:setup-compute}

All training and evaluation are run on a GPU cluster using NVIDIA H100 (80\,GB) and H200 (141\,GB) GPUs. We use one GPU per run; multi-node training was \emph{not} required for any reported result.

\begin{itemize}
\item \textbf{DNA/RNA CPT}: $1\!\times$ H100, walltime up to $3$~days per run. Eight runs $\approx 24$ GPU-days.
\item \textbf{Protein SFT}: $1\!\times$ H100, walltime up to $3$~days per run. Eight runs across the reported data and epoch sweeps.
\item \textbf{Protein RL}: $1\!\times$ H200, walltime up to $7$~days per run with automatic checkpointing and requeueing. The reported warm-start $\beta=10^{-4}$ sweep comprises $8$ runs, corresponding to the protein RL results in Figure~\ref{fig:rl_epoch_scaling}. Each training run chains two GPU evaluation jobs, one for the ID split and one for the OOD split, with sampling at $T=0.7$. We did not include the exploratory $\beta=10^{-3}$ and no-warm-start protein RL runs in the reported results, so they are excluded from the compute accounting here.
\item \textbf{DNA/RNA SFT/RL}: $1\!\times$ H100, walltime up to $7$~days per run. SFT and RL are submitted as chained pairs per backbone, with RL automatically selecting the best SFT checkpoint by validation accuracy.
\end{itemize}

The dominant reported compute cost is the protein RL sweep: $8$ training runs at up to $7$~days each (${\approx}56$ H200-days), plus $16$ paired ID/OOD evaluation jobs (${\approx}20$ GPU-days). The DNA/RNA CPT and protein SFT sweeps consume the next-largest shares at ${\approx}24$ GPU-days each.

\section{Main-Results Tables}

\subsection{Results for Figure 2}

\begin{table}[h]
\centering
\resizebox{\textwidth}{!}{
\begin{tabular}{lllrrrrrrr}
\toprule
Task & Model & Split & Base & SFT1 & SFT2 & SFT4 & SFT8 & SFT16 & SFT32 \\
\midrule
DNA & Qwen3-1.7B & ID  & 0.562 & 0.724 & 0.838 & 0.866 & 0.893 & 0.907 & 0.879 \\
DNA & Qwen3-1.7B & OOD & 0.576 & 0.681 & 0.714 & 0.736 & 0.725 & 0.703 & 0.687 \\
DNA & Qwen3-4B   & ID  & 0.593 & 0.848 & 0.886 & 0.890 & 0.838 & 0.845 & 0.852 \\
DNA & Qwen3-4B   & OOD & 0.582 & 0.702 & 0.736 & 0.755 & 0.752 & 0.734 & 0.710 \\
\midrule
RNA & Qwen3-1.7B & ID  & 0.219 & 0.614 & 0.861 & 0.875 & 0.861 & 0.875 & 0.861 \\
RNA & Qwen3-1.7B & OOD & 0.195 & 0.365 & 0.547 & 0.547 & 0.568 & 0.561 & 0.541 \\
RNA & Qwen3-4B   & ID  & 0.226 & 0.778 & 0.847 & 0.892 & 0.903 & 0.911 & 0.886 \\
RNA & Qwen3-4B   & OOD & 0.202 & 0.543 & 0.584 & 0.606 & 0.627 & 0.594 & 0.569 \\
\midrule
Protein & Qwen3-1.7B & ID  & 0.103 & 0.282 & 0.330 & 0.357 & 0.368 & 0.333 & 0.315 \\
Protein & Qwen3-1.7B & OOD & 0.095 & 0.220 & 0.254 & 0.278 & 0.261 & 0.245 & 0.233 \\
Protein & Qwen3-4B   & ID  & 0.126 & 0.303 & 0.349 & 0.368 & 0.382 & 0.367 & 0.332 \\
Protein & Qwen3-4B   & OOD & 0.108 & 0.322 & 0.364 & 0.340 & 0.342 & 0.305 & 0.291 \\
\bottomrule
\end{tabular}
}
\caption{Numerical results corresponding to Figure~2. DNA and RNA are evaluated with accuracy; proteins are evaluated with propagated unweighted $F_{\max}$. Values are reported for the base model at epoch 0 and after supervised fine-tuning for the indicated number of epochs.}
\label{tab:fig2_full_results}
\end{table}

\newpage

\subsection{Results for Figure 3}

\begin{table}[h]
\centering
\begin{tabular}{lllrrrrrr}
\toprule
Task & Model & Split & Base & 4K & 8K & 12K & 16K & 20K \\
\midrule
Protein & Qwen3-1.7B & ID  & 0.102 & 0.178 & 0.214 & 0.231 & 0.245 & 0.250 \\
Protein & Qwen3-1.7B & OOD & 0.095 & 0.122 & 0.188 & 0.218 & 0.216 & 0.222 \\
Protein & Qwen3-4B   & ID  & 0.118 & 0.222 & 0.266 & 0.248 & 0.278 & 0.269 \\
Protein & Qwen3-4B   & OOD & 0.110 & 0.237 & 0.291 & 0.289 & 0.314 & 0.308 \\
\bottomrule
\end{tabular}
\caption{Full numerical results corresponding to Figure~3 for protein function prediction under fixed-compute, variable-data supervised fine-tuning. All post-training runs use one SFT epoch while varying the number of training examples. Metrics are propagated unweighted $F_{\max}$ on the ID and OOD test splits. Values are reported for the base model at 0K and after SFT on the indicated number of training examples.}
\label{tab:fig3_full_results}
\end{table}

\subsection{Results for Figure 4}

\begin{table}[h]
\centering
\resizebox{\textwidth}{!}{
\begin{tabular}{lllrrrrr}
\toprule
Task & Model & Split & RL1 & RL2 & RL4 & RL8 & RL16 \\
\midrule
DNA & Qwen3-1.7B & ID  & 0.860 & 0.939 & 0.932 & 0.930 & 0.924 \\
DNA & Qwen3-1.7B & OOD & 0.720 & 0.751 & 0.782 & 0.793 & 0.815 \\
DNA & Qwen3-4B   & ID  & 0.930 & 0.980 & 0.971 & 0.946 & 0.952 \\
DNA & Qwen3-4B   & OOD & 0.738 & 0.783 & 0.805 & 0.818 & 0.838 \\
\midrule
RNA & Qwen3-1.7B & ID  & 0.625 & 0.722 & 0.736 & 0.777 & 0.785 \\
RNA & Qwen3-1.7B & OOD & 0.500 & 0.655 & 0.669 & 0.642 & 0.689 \\
RNA & Qwen3-4B   & ID  & 0.775 & 0.847 & 0.893 & 0.902 & 0.914 \\
RNA & Qwen3-4B   & OOD & 0.582 & 0.694 & 0.708 & 0.732 & 0.745 \\
\midrule
Protein & Qwen3-1.7B & ID  & 0.648 & 0.776 & 0.797 & 0.805 & -- \\
Protein & Qwen3-1.7B & OOD & 0.592 & 0.710 & 0.734 & 0.738 & -- \\
Protein & Qwen3-4B   & ID  & 0.697 & 0.870 & 0.930 & 0.899 & -- \\
Protein & Qwen3-4B   & OOD & 0.682 & 0.893 & 0.956 & 0.909 & -- \\
\bottomrule
\end{tabular}
}
\caption{Full numerical results corresponding to Figure~4. DNA and RNA are evaluated with accuracy; proteins are evaluated with propagated unweighted $F_{\max}$. Columns report performance after the indicated number of reinforcement-learning epochs. Protein RL was evaluated through 8 epochs in this figure.}
\label{tab:fig4_full_results}
\end{table}

\subsection{Results for Figure 5}

\begin{table}[h]
\centering
\resizebox{\textwidth}{!}{
\begin{tabular}{lllrrrrr}
\toprule
Task & Model & Split & Base & SFT & SFT+RL & CPT+SFT & CPT+SFT+RL \\
\midrule
DNA & Qwen3-1.7B & ID  & 0.721 & 0.905 & 0.939 & 0.835 & 0.965 \\
DNA & Qwen3-1.7B & OOD & 0.651 & 0.724 & 0.893 & 0.935 & 0.959 \\
DNA & Qwen3-4B   & ID  & 0.798 & 0.894 & 0.980 & 0.917 & 0.986 \\
DNA & Qwen3-4B   & OOD & 0.668 & 0.753 & 0.917 & 0.924 & 0.970 \\
\midrule
RNA & Qwen3-1.7B & ID  & 0.209 & 0.890 & 0.773 & 0.902 & 0.936 \\
RNA & Qwen3-1.7B & OOD & 0.181 & 0.569 & 0.692 & 0.718 & 0.754 \\
RNA & Qwen3-4B   & ID  & 0.218 & 0.912 & 0.918 & 0.910 & 0.939 \\
RNA & Qwen3-4B   & OOD & 0.197 & 0.588 & 0.744 & 0.732 & 0.765 \\
\bottomrule
\end{tabular}
}
\caption{Full numerical results corresponding to Figure~5 for the continued-pretraining ablation. DNA and RNA are evaluated with accuracy on the ID and OOD test splits. The Base column reports the non-CPT backbone before task-specific post-training. SFT and SFT+RL report the strongest non-CPT post-training configurations, while CPT+SFT and CPT+SFT+RL report the corresponding configurations initialized from the CPT-adapted backbone.}
\label{tab:fig5_full_results}
\end{table}

\newpage

\subsection{Results for Figure 6}

\begin{table}[h]
\centering
\resizebox{\textwidth}{!}{
\begin{tabular}{llrrrrrrr}
\toprule
Stage & Split & Base & Epoch 1 & Epoch 2 & Epoch 4 & Epoch 8 & Epoch 16 & Epoch 32 \\
\midrule
SFT & ID  & 0.223 & 0.778 & 0.750 & 0.806 & 0.778 & 0.815 & 0.815 \\
SFT & OOD & 0.206 & 0.520 & 0.534 & 0.547 & 0.561 & 0.564 & 0.564 \\
\midrule
RL  & ID  & 0.815 & 0.826 & 0.853 & 0.882 & 0.890 & 0.892 & -- \\
RL  & OOD & 0.564 & 0.599 & 0.684 & 0.702 & 0.724 & 0.729 & -- \\
\bottomrule
\end{tabular}
}
\caption{Full numerical results corresponding to Figure~6 for the Gemma4-E2B RNA backbone ablation. Metrics are accuracy on the RNA ID and OOD test splits. The SFT rows report the supervised fine-tuning epoch sweep, including the base model at epoch 0. The RL rows report the reinforcement-learning epoch sweep initialized from the strongest SFT checkpoint; RL was evaluated through 16 epochs.}
\label{tab:fig6_gemma_full_results}
\end{table}

\subsection{Results for Figure 8}

\begin{table}[h]
\centering
\resizebox{\textwidth}{!}{
\begin{tabular}{lllrrrrrrrrr}
\toprule
Task & Model & Split & SFT0/RL8 & SFT1/RL7 & SFT2/RL6 & SFT3/RL5 & SFT4/RL4 & SFT5/RL3 & SFT6/RL2 & SFT7/RL1 & SFT8/RL0 \\
\midrule
DNA & Qwen3-1.7B & ID  & 0.842 & 0.899 & 0.937 & 0.929 & 0.920 & 0.879 & 0.843 & 0.844 & 0.891 \\
DNA & Qwen3-1.7B & OOD & 0.618 & 0.746 & 0.752 & 0.781 & 0.752 & 0.731 & 0.701 & 0.715 & 0.713 \\
DNA & Gemma4-E2B & ID  & 0.864 & 0.891 & 0.927 & 0.946 & 0.934 & 0.918 & 0.903 & 0.888 & 0.865 \\
DNA & Gemma4-E2B & OOD & 0.685 & 0.762 & 0.796 & 0.785 & 0.772 & 0.749 & 0.713 & 0.693 & 0.676 \\
\midrule
RNA & Qwen3-1.7B & ID  & 0.722 & 0.819 & 0.847 & 0.819 & 0.736 & 0.639 & 0.653 & 0.694 & 0.861 \\
RNA & Qwen3-1.7B & OOD & 0.662 & 0.777 & 0.757 & 0.770 & 0.669 & 0.566 & 0.549 & 0.554 & 0.568 \\
RNA & Gemma4-E2B & ID  & 0.754 & 0.839 & 0.862 & 0.876 & 0.884 & 0.842 & 0.817 & 0.791 & 0.776 \\
RNA & Gemma4-E2B & OOD & 0.683 & 0.726 & 0.735 & 0.748 & 0.714 & 0.665 & 0.638 & 0.591 & 0.576 \\
\bottomrule
\end{tabular}
}
\caption{Full numerical results corresponding to Figure~8 for the fixed-budget SFT--RL allocation experiment. DNA and RNA are evaluated with accuracy on the ID and OOD test splits. The total post-training budget is fixed to eight epoch-level passes, and columns indicate the allocation between supervised fine-tuning and reinforcement learning. Values are reported as proportions.}
\label{tab:fig8_full_results}
\end{table}

\section{Additional Experiments}

\subsection{Scaling Post-Training for Biological Non-Reasoning Tasks}
\label{app:non_reasoning_sft_scaling}

The main experiments in this paper focus on biological reasoning tasks, where the model must integrate biological inputs with natural-language context and produce mechanistic or structured outputs. To test whether the same post-training trends also appear in a more conventional biological prediction setting, we additionally evaluate supervised fine-tuning on variant effect prediction (VEP), using the coding non-SNV benchmark introduced in \citep{fallahpour2025bioreason}. Unlike the KEGG-derived pathway prediction task, which requires multi-step mechanistic inference over molecular networks, VEP-Non-SNV is primarily a classification-style task: given paired reference and variant DNA sequences together with gene and chromosome context, the model predicts whether a coding non-SNV is benign or pathogenic, and, when pathogenic, the associated disease phenotype.

The VEP-Non-SNV dataset is constructed from ClinVar coding non-SNVs, filtered to include nuclear-genome variants affecting at most 64 base pairs, with sufficient clinical review status and transcript matching to GRCh38.p14. The original benchmark in \citep{fallahpour2025bioreason} uses stratified train/test partitioning to balance disease representation and augments each example with paraphrased prompts. This makes the task biologically meaningful, but less dependent on explicit chain-of-thought-style mechanistic reasoning than the pathway prediction benchmark. In our setting, this experiment therefore serves as a non-reasoning control for testing whether increasing SFT compute continues to improve performance monotonically. \citep{fallahpour2025bioreason} describes this benchmark as containing 36,088 core non-SNV entries and defines the task as predicting benign versus pathogenic status, with conditional disease prediction for pathogenic variants.

\begin{table}[h]
\centering
\small
\begin{tabular}{lcccccc}
\toprule
SFT Epochs & 1 & 2 & 4 & 8 & 16 & 32 \\
\midrule
Accuracy & 0.7123 & 0.7789 & 0.7965 & 0.8246 & \textbf{0.8316} & 0.8105 \\
\bottomrule
\end{tabular}
\caption{
SFT epoch scaling on the VEP-Non-SNV task. The model improves with additional supervised fine-tuning up to 16 epochs, after which performance declines, suggesting that even non-reasoning biological prediction tasks can exhibit non-monotonic SFT scaling.
}
\label{tab:vep_non_snv_sft_scaling}
\end{table}

Table~\ref{tab:vep_non_snv_sft_scaling} reports SFT epoch scaling accuracy for Qwen3-1.7B on VEP-Non-SNV. Performance improves steadily from one to sixteen epochs, increasing from 0.7123 at one epoch to 0.8316 at sixteen epochs, before declining at thirty-two epochs. Thus, even in this less explicitly reasoning-oriented setting, scaling SFT is not strictly monotonic: moderate additional supervision improves performance, but excessive training begins to degrade the final metric. This mirrors the broader pattern observed in the main biological reasoning experiments, where SFT is a strong driver of task performance but can over-specialize when scaled too far.

\subsection{DNA LoRA Rank Allocation}
\label{app:dna_lora}

\begin{figure*}[h!]
    \centering
    \vspace{-5mm}
    \includegraphics[width=\textwidth]{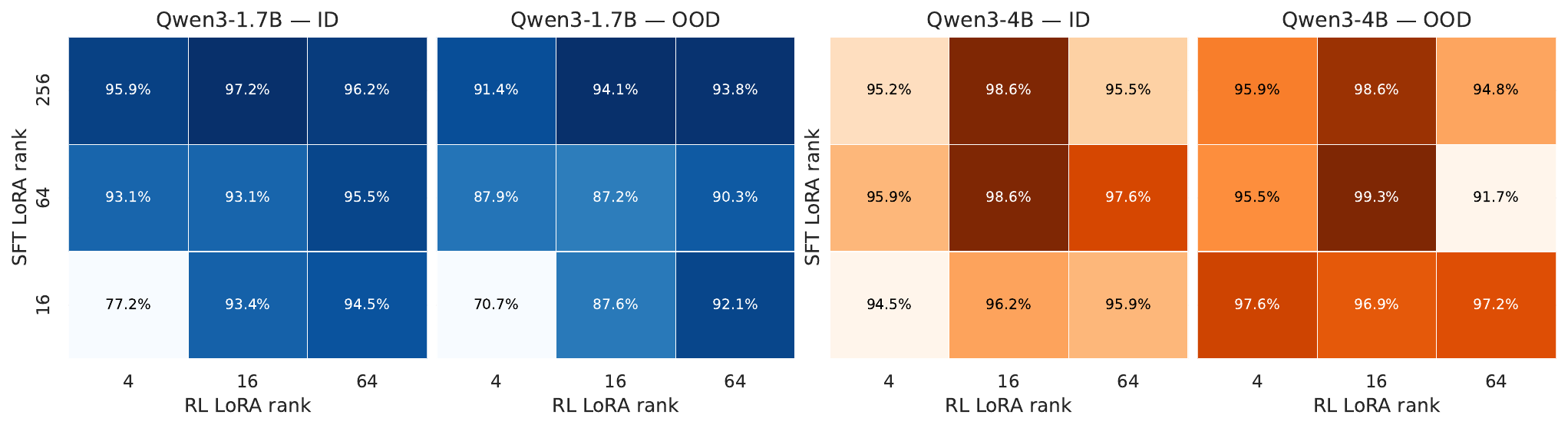}
    \vspace{-5mm}
    \caption{\textbf{Optimal adaptation requires asymmetric capacity across training stages.} Higher LoRA rank benefits SFT, while lower rank is sufficient for RL, indicating that different stages require different adaptation capacity (both for ID and OOD tasks). Shown are results for pathway prediction (DNA) tasks.
    }
    \label{fig:dna_lora}
    \vspace{-0.8\baselineskip}
\end{figure*}

\subsection{Fixed Epoch, Variable Data during RL}
\label{app:rl_data_scaling}

In the main text, we demonstrated non-monotonic behavior when scaling RL \emph{epochs} at fixed data size. A complementary question is whether scaling the \emph{number of RL examples} at a fixed single epoch yields similar saturation. Table~\ref{tab:protein-rl-klstrong} reports protein function prediction performance (propagated GO-F1) for Qwen3-4B-Thinking trained with GRPO at $\beta=10^{-3}$ (strong KL) for one epoch, varying the number of training examples from 4K to 20K.

Both ID and OOD F1 peak at an intermediate data budget (12K for ID, 4K for OOD) and decline as additional examples are added. The best OOD F1 (0.956 at 4K) exceeds the best ID F1 (0.952 at 12K), consistent with the observation in the main text that moderate RL improves generalization disproportionately. At the largest budget (20K), both metrics drop substantially below their respective peaks, with OOD F1 falling from 0.956 to 0.884. This indicates that, under a fixed single-epoch schedule, additional RL data does not substitute for the exploration benefits of multi-epoch training and can instead over-constrain the policy. The result complements the epoch-scaling findings and reinforces the broader conclusion that RL compute allocation---whether measured in epochs or data volume---requires careful calibration to maximize impact on biological reasoning capabilities.

\begin{table}[ht]
\centering
\begin{tabular}{lccccc}
\toprule
\textbf{Metric} & \textbf{4K} & \textbf{8K} & \textbf{12K} & \textbf{16K} & \textbf{20K} \\
\midrule
ID F1  & $0.935$ & $0.931$ & $\mathbf{0.952}$ & $0.901$ & $0.877$ \\
OOD F1 & $\mathbf{0.956}$ & $0.954$ & $0.938$ & $0.921$ & $0.884$ \\
\bottomrule
\end{tabular}
\caption{Protein function prediction F1 (Qwen3-4B-Thinking, GRPO $\beta=10^{-3}$, 1 epoch) as a function of RL training examples. Increasing data does not monotonically improve performance; both ID and OOD F1 peak at intermediate budgets and decline with further scaling.}
\label{tab:protein-rl-klstrong}
\end{table}

\subsection{Fixed Budget, Variable Data Allocation between SFT and RL}
\label{app:data-allocation-sft-rl}

We next ask whether the fixed-budget SFT--RL trade-off from Section~4.6 also appears when the
total number of post-training examples is fixed and only the data allocation between SFT and RL is
varied. Table~\ref{tab:data-allocation-sft-rl} reports protein function prediction performance under
a fixed 20K-example post-training budget. Each row allocates a different fraction of the data to SFT
and RL while keeping the total number of examples constant.

\begin{table}[h]
\centering
\small
\setlength{\tabcolsep}{6pt}
\renewcommand{\arraystretch}{1.12}
\begin{tabular}{@{}lrrrr@{}}
\toprule
\textbf{SFT / RL data} & \textbf{SFT $n$} & \textbf{RL $n$} & \textbf{ID F1} & \textbf{OOD F1} \\
\midrule
0\% / 100\%   & 0     & 20000 & 0.7711          & 0.7953          \\
20\% / 80\%  & 4000  & 16000 & \textbf{0.9470} & \textbf{0.9685} \\
25\% / 75\%  & 5000  & 15000 & 0.9432          & 0.9620          \\
40\% / 60\%  & 8000  & 12000 & 0.9167          & 0.9360          \\
50\% / 50\%  & 10000 & 10000 & 0.9285          & 0.9569          \\
60\% / 40\%  & 12000 & 8000  & 0.9055          & 0.9102          \\
80\% / 20\%  & 16000 & 4000  & 0.8816          & 0.9065          \\
100\% / 0\%  & 20000 & 0     & 0.2768          & 0.3177          \\
\bottomrule
\end{tabular}
\caption{Fixed-budget data allocation between supervised fine-tuning and reinforcement learning. 
The total post-training data budget is fixed at 20K examples, and rows vary the fraction assigned to 
SFT versus RL. Metrics are reported as F1 on the ID and OOD protein function prediction splits.}
\label{tab:data-allocation-sft-rl}
\end{table}

The best results occur in the mixed SFT--RL regime rather than at either endpoint. Allocating a small
fraction of examples to SFT and the majority to RL gives the strongest ID and OOD performance,
with the 20\%/80\% split reaching 0.9470 ID F1 and 0.9685 OOD F1. Pure SFT performs poorly in
this setting, while pure RL is substantially better but still far below the mixed allocations, indicating
that RL benefits from a modest supervised warm start even when the total data budget is fixed.

\subsection{Reward Model Ablations}

\begin{table}[h]
\centering
\resizebox{\textwidth}{!}{%
\begin{tabular}{lllrrrrrrr}
\toprule
\textbf{Domain} &
\textbf{Model} &
\textbf{SFT init.} &
\textbf{RL epoch} &
\textbf{OOD Acc.} &
\textbf{Format-valid \%} &
\textbf{Parseable final \%} &
\textbf{Format-only \%} &
\textbf{Empty/no-answer \%} &
\textbf{Mean tokens} \\
\midrule

RNA & Qwen3-1.7B & S8 & 0 & 0.567 & 100.0 & 69.6  & 43.2 & 0.0 & 201.1 \\
RNA & Qwen3-1.7B & S4 & 0 & 0.547 & 100.0 & 79.1  & 45.3 & 0.0 & 210.5 \\
RNA & Qwen3-1.7B & S4 & 1 & 0.500 & 100.0 & 100.0 & 50.0 & 0.0 & 568.2 \\
RNA & Qwen3-1.7B & S4 & 2 & 0.655 & 100.0 & 100.0 & 34.5 & 0.0 & 170.9 \\
RNA & Qwen3-1.7B & S4 & 4 & 0.669 & 100.0 & 100.0 & 33.1 & 0.0 & 419.6 \\
RNA & Qwen3-1.7B & S4 & 8 & 0.642 & 100.0 & 100.0 & 35.8 & 0.0 & 544.6 \\

\bottomrule
\end{tabular}%
}
\caption{OOD reward-hacking audit across RNA RL checkpoints. Epoch $0$ denotes the SFT initialization before RL. Format-only success is defined as a format-valid output with an incorrect final answer.}
\label{tab:reward-hacking-audit}
\end{table}

\subsection{ID/ OOD Split Ablations}
\label{app:rna-celltype-split}

To test whether the qualitative training dynamics depend on how we define in-domain and out-of-domain settings, we provide an additional ablation in the RNA setting. Instead of splitting the data by held-out disease, we construct an alternative cell-type split for the target-identification task. We label an example as OOD if its canonical cell type is one of \emph{regulatory T cell}, \emph{exhausted T cell}, or \emph{myeloid cell}, and use the remaining examples for training and validation. This yields 1,418 training examples, 75 validation examples, and 102 OOD test examples. The task format, model architecture, and evaluation protocol are otherwise unchanged from the main RNA experiments: the model receives the disease, cell type, five candidate genes, and aligned TranscriptFormer representations, and is evaluated by greedy generation with exact match against the target gene. This split therefore changes only the biological axis of distribution shift, from held-out disease to held-out cellular context, while preserving the same target-identification setup described earlier.

\begin{table}[h]
\centering
\begin{tabular}{llcc}
\toprule
\textbf{Stage} & \textbf{Checkpoint} & \textbf{ID val.} & \textbf{OOD test} \\
\midrule
SFT & 1 epoch & 54.7 & 60.8 \\
SFT & 2 epochs & 54.7 & 43.1 \\
SFT & 4 epochs & 58.7 & 39.2 \\
SFT & 8 epochs & 62.7 & 41.2 \\
\midrule
RL & 1 epoch & 65.3 & 88.2 \\
RL & 2 epochs & 68.0 & 92.2 \\
RL & 4 epochs & 66.7 & 93.1 \\
RL & 8 epochs & 73.3 & 95.1 \\
\bottomrule
\end{tabular}
\caption{RNA target-identification performance under the held-out cell-type split. OOD examples are those whose canonical cell type is \emph{regulatory T cell}, \emph{exhausted T cell}, or \emph{myeloid cell}. All values are exact-match accuracies, reported as percentages, under greedy generation.}
\label{tab:rna-celltype-split}
\end{table}

Under this cell-type split, supervised fine-tuning again improves in-domain performance while failing to improve OOD generalization. Qwen3-1.7B reaches 54.7\% ID accuracy after one epoch and increases to 62.7\% by eight epochs, but OOD accuracy drops from 60.8\% at one epoch to 39.2--43.1\% across later SFT checkpoints. Starting GRPO from the four-epoch SFT checkpoint reverses this behavior: OOD accuracy rises from 39.2\% at initialization to 88.2\% after one RL epoch and continues improving to 95.1\% after eight RL epochs, while ID accuracy also increases from 58.7\% to 73.3\%. This ablation supports the main conclusion that the SFT--RL contrast is not specific to the held-out hepatoblastoma disease split used in the main RNA experiments. Instead, the same qualitative pattern appears under a distinct biologically meaningful OOD definition: SFT fits the training/validation distribution, whereas RL substantially improves transfer to held-out cellular contexts.

\newpage

\section{Asset Licenses and Redistribution Status}

\begin{table}[h]
\centering
\caption{Existing assets used in this work and their licenses or terms of use.}
\label{tab:assets-licenses}
\resizebox{\textwidth}{!}{%
\begin{tabular}{p{0.24\textwidth} p{0.30\textwidth} p{0.41\textwidth}}
\toprule
\textbf{Asset} & \textbf{Used in this paper} & \textbf{License or terms of use} \\
\midrule

BIOREASON pathway prediction benchmark (\texttt{wanglab/kegg}) &
Pathway prediction benchmark for DNA reasoning experiments &
Apache-2.0, as listed on the Hugging Face dataset card. \\

MEDEA / MedeaDB target-identification task &
RNA target-identification benchmark and disease/cell-type/candidate-gene setup. &
The MEDEA code repository is Apache-2.0. The benchmark data are distributed separately
through the mims-harvard/MedeaDB Hugging Face dataset (CC BY-NC-SA 4.0) referenced by the MEDEA README.
We use the targetID task structure and TranscriptFormer setting described in MEDEA.
We do not redistribute the raw MedeaDB data unless permitted by the MedeaDB dataset
license/terms; any derived splits or preprocessing scripts will be released only under
terms compatible with the upstream data license. \\

BioReason-Pro SFT reasoning data (\texttt{wanglab/bioreason-pro-sft-reasoning-data}) &
Protein function reasoning benchmark/data &
Apache-2.0, as listed on the Hugging Face dataset card. \\

FineFineWeb biology subset &
Continued pre-training corpus &
Open Data Commons Attribution License, ODC-By v1.0; use is also subject to CommonCrawl Terms of Use. \\

Qwen3-1.7B / Qwen3-4B &
Text LLM backbones for DNA, RNA, and protein experiments &
Apache-2.0, as listed for Qwen3 model files on Hugging Face. \\

Gemma 4 E2B &
Backbone ablation for RNA experiments &
Apache-2.0, as stated in the official Gemma 4 model card. \\

Evo2-1B &
Frozen DNA encoder &
Apache-2.0 for the Evo2 repository and released model artifacts. \\

TranscriptFormer &
Frozen transcriptomic encoder &
MIT License, as listed by the TranscriptFormer repository/model page. \\

ESM-3 / ESM3-1B protein encoder &
Frozen protein encoder &
Custom non-commercial model-weight terms for the ESM-3 Open Model checkpoint used, including EvolutionaryScale's Cambrian Non-Commercial License Agreement where applicable. \\

\bottomrule
\end{tabular}%
}
\end{table}




\end{document}